\documentclass[3p]{elsarticle}

\usepackage{natbib}
\usepackage{amsmath,amssymb,amsfonts}
\usepackage{graphicx}
\usepackage{textcomp}
\usepackage{xcolor}
\usepackage{hyperref}
\usepackage{amsbsy}
\usepackage{amsthm}
\usepackage{algorithmic}
\usepackage[]{algorithm}
\usepackage{mathtools}
\usepackage[toc,page,header]{appendix}
\usepackage{minitoc}
\usepackage[thinc]{esdiff}
\usepackage{stackengine}
\usepackage{appendix}
\usepackage{multirow}
\usepackage{lipsum} 
\usepackage{tikz}
\usepackage{amssymb}

\providecommand{\abs}[1]{\lvert#1\rvert}
\newcommand{\norm}[1]{\ensuremath{\left\|#1\right\|}} 
\newcommand{\leqtext}[1]{\ensuremath{\stackrel{\text{#1}}{\leq}}} 
\newcommand{\geqtext}[1]{\ensuremath{\stackrel{\text{#1}}{\geq}}} 
\newcommand{\preceqtext}[1]{\ensuremath{\stackrel{\text{#1}}{\preceq}}}

\DeclareMathOperator*{\argmin}{arg\,min} 
\providecommand{\ip}[2]{\langle #1, #2 \rangle} 

\DeclarePairedDelimiter\ceil{\lceil}{\rceil}

\providecommand{\ip}[2]{\langle #1, #2 \rangle} 
\newcommand{\px}[2]{\ensuremath{\mathcal{P}_{#1}\left(#2\right)}} 

\def \x {{\boldsymbol{x}}}
\def \y {{\boldsymbol{y}}}
\def \z {{\boldsymbol{z}}}

\def \u {{\boldsymbol{u}}}
\def \v {{\boldsymbol{v}}}

\def \bx {{\boldsymbol{x}}}
\def \bu {{\boldsymbol{u}}}

\def\opt{\mathsf{OPT}}
\def \bbR{{\mathbb{R}}}
\def \bX{{\mathcal{X}}}

\def \A {{\boldsymbol{A}}}
\def \B {{\boldsymbol{B}}}
\def \C {{\boldsymbol{C}}}
\def \H {{\boldsymbol{H}}}
\def \I {{\boldsymbol{I}}}

\def \P {{\boldsymbol{P}}}

\def \X {{\boldsymbol{X}}}
\def \th {{\boldsymbol{\theta}}}

\def \T {{\mathsf{T}}}
\def \P {{\mathcal{P}}}

\def \cA {{\c{A}}}
\def \cB {{\c{B}}}

\def \cX {{\c{X}}}

\def \Rn {{\mathbb{R}}}

\def\cA{\mathcal{A}}
\def\cB{\mathcal{B}}

\def\cO{\mathcal{O}}
\def\cP{\mathcal{P}}

\def\cX{\mathcal{X}}

\def\nn{\nonumber}

\newtheorem{assumption}{}

\theoremstyle{remark}
\newtheorem{rem}{\bf Remark}
\newtheorem{theorem}{Theorem}
\newtheorem{lemma}{Lemma}

\newtheorem{definition}{Definition}

\newtheorem{corollary}{Corollary}

\makeatletter
\def\blfootnote{\gdef\@thefnmark{}\@footnotetext}
\makeatother

\begin{document}
	\begin{frontmatter}
		\title{Online Convex Optimization with Switching Cost and Delayed Gradients}
		\author[spandan]{Spandan Senapati}
		\ead{spandansenapatiphy@gmail.com}
		\author[rahul]{Rahul Vaze}
		\ead{rahul.vaze@gmail.com}
		\address[spandan]{Department of Computer Science and Engineering, Indian Institute of Technology Kanpur, Kanpur, India}
		\address[rahul]{School of Technology and Computer Science, Tata Institute of Fundamental Research, Mumbai, India}		
		\date{}
		\begin{abstract}%
			We consider the \textit{online convex optimization (OCO)} problem with \textit{quadratic} and \textit{linear} switching cost in the \textit{limited information} setting, where an online algorithm can choose its action using only gradient information about the previous objective function. For $L$-smooth and $\mu$-strongly convex objective functions, we propose an online multiple gradient descent (OMGD) algorithm and show that its competitive ratio for the OCO problem with quadratic switching cost is at most $4(L + 5) + \frac{16(L + 5)}{\mu}$. The competitive ratio upper bound for OMGD is also shown to be order-wise tight in terms of $L,\mu$. In addition, we show that the competitive ratio of any online algorithm is $\max\{\Omega(L), \Omega(\frac{L}{\sqrt{\mu}})\}$ in the limited information setting when the switching cost is quadratic. We also show that the OMGD algorithm achieves the optimal (order-wise) dynamic regret in the limited information setting.
			For the linear switching cost, the competitive ratio upper bound of the OMGD algorithm is shown to depend on both the path length and the squared path length of the problem instance, in addition to \(L, \mu\), and is shown to be order-wise, the best competitive ratio any online algorithm can achieve. Consequently, we conclude that the optimal competitive ratio for the quadratic and linear switching costs are fundamentally different in the limited information setting.
			

		\end{abstract}
	\end{frontmatter}
	\section{Introduction}\label{sec:introduction}
Online convex optimization (OCO) is a very popular optimization paradigm where decisions have to be made with  unknown utility functions. 
With OCO, at each time $t$,
the goal of an online algorithm $\cA$ is to choose an action $\boldsymbol{x}_t$ that minimizes a  convex objective function $f_t$, however $f_t$ is revealed after $\x_t \in \cX \subseteq \Rn^ d$ has been chosen. In most settings with OCO, once $\x_t$ is chosen both $f_t(\x_t)$ and $\nabla f_t(\x_t)$ are revealed. 
Knowing all the $f_t$'s, $t = 1, \dots, T$ ahead of time, where $T$ is the time horizon, an optimal offline algorithm ($\opt$) chooses action $\x^\star := \arg \min_{\x \in \cX} \sum_{t=1}^T f_t(\x)$, and the {\it static} regret for $\cA$ is defined as 
$Sr \coloneqq  \max_{f_t, t=1, \dots, T}\sum_{t=1}^T f_t(\x_t) - \sum_{t=1}^T f_t(\x^\star),$ i.e., an adversary can choose the functions $f_t$. The name static comes from the 
fact that $\opt$ is constrained to use a single action. 

Large body of work is known for  static regret \cite{srebro2010smoothness, bartlett2009adaptive, garber2020online, hazan2019introduction,shalev2011online}, where if the functions $f_t$'s are convex, the optimal regret is $\Theta(\sqrt{T})$, while if the functions $f_t$'s are strongly convex
then algorithms with regret at most $\mathcal{O}(\log T)$ are known. When the functions $f_t$'s are chosen 
by an adversary, but the arrival order is randomized, algorithms with better guarantees are also
known \cite{garber2020online}.

Natural generalization of the static regret is the dynamic regret \cite{zinkevich2003online,hall2013dynamical}, where the regret for a comparator sequence $\bu = (\u_1, \dots, \u_T)$ is defined as 
\(
Dr(\bu) := \max_{f_t, t=1, \dots, T}\sum_{t=1}^T f_t(\x_t) - \sum_{t=1}^T f_t(\u_t). 
\)
For this general dynamic regret definition, sub-linear (in $T$) regret is not always possible, unless some structure is enforced on the comparator sequence. 
For example, for a sequence $\bu$, defining $\cP_{T}(\u) := \sum_{t=2}^T ||\u_t - \u_{t-1}||$,
the online gradient descent (OGD) algorithm was shown to achieve dynamic regret $Dr(\bu) = \mathcal{O}(\sqrt{T}(1+\cP_T(\u))$ \cite{zinkevich2003online}, which has been improved to 
$\mathcal{O}(\sqrt{T(1+\cP_T(\u))})$ in \cite{zhang2018adaptive}, matching the lower bound $\Omega(\sqrt{T(1+\cP_T(\u))})$ \cite{zhang2018adaptive}. 

A special case of \(Dr(\u)\) that is popularly studied \cite{besbes2015non, jadbabaie2015online, yang2016tracking, cdcstrictconvex, zhang2017improved,zhao2020bandit, zhang2020minimizing, zhao2021improved} is by restricting $\bu =  \bx^\star= (\x_1^\star, \dots, \x_T^\star)$, where $\x_t^\star \coloneqq \arg \min_{\x \in \cX}  f_t(\x)$ is the local minimizer of \(f_t(\x)\). 
For this case, the best known bound on the dynamic regret has been shown to be $\mathcal{O}(\cP_T(\bx^\star))$ in \cite{cdcstrictconvex} by using the online gradient descent (OGD) algorithm, when the functions $f_t$'s are strongly convex and smooth. 
Under the special case that the minimizers $\x_t^\star$ lie in the interior of the feasible set, $\mathcal{O}(\cP_T(\bx^\star))$ regret can be 
achieved when the functions $f_t$'s are just convex and smooth \cite{yang2016tracking}. 
For strongly convex and smooth functions, defining $\cP_{2,T}^\star :=\sum_{t=2}^T ||\x_t^\star - \x_{t-1}^\star||^2$, \cite{zhang2017improved} showed that $\mathcal{O}(\min\{\cP_T(\bx^\star), \cP_{2,T}^\star\})$ is also 
achievable as long as at each time $t$, the gradient of $f_t$ at multiple points is available. Most recently, \cite{zhao2021improved} improved this guarantee to 
$\mathcal{O}(\min\{\cP_T(\bx^\star), \cP_{2,T}^\star, FV\})$, where 
$FV :=  \sum_{t=2}^T\sup_{\x \in \mathcal{X}} |f_t(\x) - f_{t-1}(\x)|$ denotes the functional variation.

The constrained version of OCO  has also been considered  recently in \cite{chen2018bandit, cao2018online, liu2022simultaneously, vaze2022dynamic}, where at each time $t$, the objective is to minimize the loss function $f_t$ subject to a constraint $g_t(\x)\le 0$. 
Compared to OCO, there are two performance measures, the usual dynamic regret and in addition the constraint violation penalty $\max_{f_t, g_t, t=1, \dots, T}\sum_{t=1}^T g_t(\x_t)$. 

One generalization of OCO  that has applications in data centers is the OCO with switching cost (OCO-S) problem, where in addition to the penalty paid for choosing sub-optimal decisions, there is a cost for changing actions in successive time slots. 
In particular, the cost of an online algorithm $\cA$ for OCO-S is 
\begin{equation}\label{cost:intro1}
	C_\cA := \sum_{t=1}^T  f_t(\bx_t) + c(\x_{t}, \x_{t - 1}),
\end{equation} 
where \(c(\x_{t}, \x_{t - 1})\) is a non-negative function that denotes the cost of switching from \(\x_{t - 1}\) to \(\x_{t}\). The goal of \(\cA\) is to  either minimize the 
dynamic regret 
\begin{equation}\label{probR:intro1}
	R_d(\cA) := \max_{f_t}(C_\cA - C_\opt),
\end{equation}
or the competitive ratio
\begin{equation}\label{prob:intro1}
	\text{cr}_\cA := \max_{f_t}\frac{C_\cA}{C_\opt},
\end{equation}
{where \(C_\opt\) denotes the cost of the $\opt$, that knows all the \(f_{t}\)'s ahead of time,  for solving \eqref{cost:intro1}}, i.e. $$C_{\opt} := \min_{\{\x_{t} \in \cX\}_{t = 1} ^ {T}} \sum_{t = 1} ^ {T}  f_t(\bx_t) + c(\x_{t}, \x_{t - 1}).$$ Two of the most common forms of the switching cost that have been considered in the literature are (a) quadratic switching cost, where \(c(\x_{t}, \x_{t - 1}) = \frac{1}{2} \norm{\x_{t} - \x_{t - 1}} ^ 2\), and (b) linear switching cost, where \(c(\x_{t}, \x_{t - 1}) = \norm{\x_{t} - \x_{t - 1}}\). In this paper, we consider both the competitive ratio \eqref{prob:intro1} as well as the dynamic regret \eqref{probR:intro1} for (a) and competitive ratio for (b).

The OCO with switching cost model is a versatile framework that captures a number of important problems in real-life. To build motivation, we consider the following problems that can be formulated within the framework. 
\begin{itemize}
	\item {Server Management in Data Centers:} The pioneering work on the OCO problem with switching cost was \cite{lin2012dynamic} that considered the problem of the dynamic right sizing for power-proportional data centers. A data center is modelled as an ensemble of servers, and it involves two important determinations: (i) determining \(x_{t}\), the count of the active servers at time \(t\), and (ii)
	 assigning jobs to servers, i.e., determining \(r_{i, t}\) which represents the arrival rate to server \(i\) at time \(t\). Let \(r_{t}\) denote the total job arrival rate at time \(t\). Then, \(\sum_{i = 1} ^ {x_{t}} r_{i, t} = r_{t}\) must hold, to ensure that all the jobs are processed.
	The goal is to choose the number of active servers \(x_{t}\) and assign jobs to each active server \(r_{i, t}\), such that the total cost incurred by the data center across \(T\) time slots is minimized.
	
	At each time \(t\), the data center's cost is modelled as a combination of (i) operational cost, representing the cost of running an active server, and (ii) the switching cost to model the cost of toggling a server on/off. The costs of operation are modelled by a convex function \(\phi(r_{i, t})\), where the function \(\phi\) is the same for all the servers. This assumption is quite standard and captures a number of commonly used server models. We refer interested readers to the discussion in \cite{lin2012dynamic} for examples of \(\phi\). The switching cost \(c(x_{t}, x_{t - 1})\) denotes the cost of changing the number of active servers. 
	
	With the cost model as mentioned above, the goal of the data center is to choose \(x_{t}\) and \(r_{i, t}\) to solve the following convex optimization problem:
	\begin{align*}
		& \min \sum_{t = 1} ^ {T} \bigg( \sum_{i = 1} ^ {x_{t}} \phi(r_{i, t}) + \beta c(x_{t}, x_{t - 1})\bigg), \\
		& \text{subject to } r_{i, t} \ge 0, \quad \sum_{i = 1} ^ {x_{t}} r_{i, t} = r_{t} \text{ for all } t,
	\end{align*}
	where the switching cost $c(x_{t}, x_{t - 1})$ can take many forms, for example $(x_{t}-x_{t - 1})^2$ or $(x_{t}-x_{t - 1})^+$.
	
	\item {Trajectory Tracking:} Given a target trajectory $\{\theta_{t}\}_{t = 1} ^ {T}$, a robot positioned at $x_{t}$ at time $t$ aims to chase the target while maintaining proximity to it, meaning the distance \(\norm{x_{t} - {\theta}_{t}}\) is small. Additionally, the robot wishes to minimize the energy expended in changing its position from \(x_{t - 1}\) to \(x_{t}\). This energy expenditure is influenced by the rate of change between positions and is proportional to the kinetic energy 
 \(\big(\frac{\norm{x_{t} - x_{t - 1}}}{\Delta}\big) ^ 2\), where $\Delta$ denotes the time taken by the robot to move from $x_{t - 1}$ to $x_{t}$. Therefore, the problem that we wish to solve for the robot can be formulated as \eqref{cost:intro1} with $f_{t}(x_{t}) = \norm{x_{t} - \theta_{t}}$ and switching cost \(c(x_{t}, x_{t - 1}) = \frac{\gamma}{2}\norm{x_{t} - x_{t - 1}} ^ 2\), or equivalently $\frac{\gamma}{2}\norm{x_{t} - x_{t - 1}}$, for a certain \(\gamma > 0\).
	
	\item {Economic Power Dispatch:} We consider the problem of economic power dispatch as outlined in \cite{li2018using}. This scenario involves a set of $k$ generators with non-negative outputs $(x_{t} ^ {(1)}, \dots, x_{t} ^ {(k)})$ (collectively represented as a vector $\x_{t} \in \mathbb{R}_{+} ^  {k}$) at a given time $t$. The main objective here is to minimize the overall generation cost $\sum_{i=1}^k g(x_{t} ^ {(i)})$, all the while ensuring a balance between the changing power demand $d_t$ and the availability of renewable energy supply $s_t$. To achieve this, a cost function $f_{t}(\x_{t})$ is introduced, which accounts for the generation cost and also incorporates a penalty for any deviation from the demand-supply equilibrium. The cost function is given by
	\begin{align}\nonumber
		f_t(\x_t) = \sum_{i=1}^k g_i(x_{t} ^ {(i)}) + \lambda_t \bigg(\sum_{i=1}^k x_{t} ^ {(i)} + s_t - d_t \bigg)^2,
	\end{align}
	where $\lambda_t > 0$ governs the penalty imposed when the demand-supply gap is not adequately maintained. Within the context of power systems, altering the power output of generators incurs significant operational costs. These are known as ramp costs and similar to \cite{li2018using, mookherjee2008dynamic, tanaka2006real}, we model the switching cost as a quadratic function $c(\x_t, \x_{t-1}) = \frac{\gamma}{2}\norm{\x_t - \x_{t-1}}^2 = \frac{\gamma}{2}\sum_{i = 1} ^ {k} (x_{t, i} - x_{t - 1, i}) ^ {2}$.
	Therefore, the goal is to minimize the total generation cost, including the imbalance penalty and the ramp costs summed across $T$ time slots. This can be formulated as \eqref{cost:intro1}  with $f_{t}(\x_{t})$ and $c(\x_{t}, \x_{t - 1})$ as discussed above.  
\end{itemize}

For the OCO-S problem with quadratic switching cost (OCO-SQ), if \(f_{t}\)'s are convex, then one can show that 
the competitive ratio \eqref{prob:intro1} of any online algorithm is unbounded. Thus, starting with \cite{chen2018}, to solve \eqref{prob:intro1}, $f_t$'s {were assumed to have properties beyond convexity, such as strong convexity.}
{The online balanced descent (OBD) algorithm, initially proposed by \cite{chen2018}, was shown to attain a competitive ratio of at most \(3 + \mathcal{O}(\frac{1}{\mu})\) for \(\mu\)-strongly convex functions in \cite{goel2019online}}. An improved variant of OBD was considered in \cite{goel2019beyond} with  competitive ratio $\mathcal{O}(\frac{1}{\sqrt{\mu}})$ as \({\mu \to 0}\), which was also shown to be the best possible (order-wise with respect to \(\mu\)) competitive ratio for all possible  online algorithms in \cite{goel2019beyond}.

OCO-S with linear switching cost (OCO-SL), where
$C_\cA = \sum_{t=1}^T  f_t(\bx_t) + \norm{\x_{t} - \x_{t - 1}}$,
has also been considered. When \(f_{t}\)'s are convex, the competitive ratio of any online algorithm has a lower bound \(\Omega(\sqrt{d})\) \cite{chen2018}. OBD  was the first algorithm to achieve a dimension free competitive ratio upper bound of \(3 + \cO(\frac{1}{\alpha})\) for \(\alpha\)-locally polyhedral functions in \cite{chen2018}. When \(f_{t}\)'s are \(\mu\)-strongly convex and \(L\)-smooth, \cite{argue2020dimension} proposed a constrained online balanced descent (COBD) algorithm with a competitive ratio \(\cO(({L / \mu}) ^ {\frac{1}{2}})\). Furthermore, \cite{argue2020dimension} showed that the competitive ratio of any online algorithm is \(\Omega(({L}/{\mu}) ^ {\frac{1}{3}})\). When $d=1$, algorithms with constant competitive ratios are known for the OCO-SL problem when $f_t$'s are just convex \cite{lin2012online, lin2012dynamic, andrew2013tale,bansal20152, albers2018optimal}.

The main limitation of most prior work on OCO-S for minimizing the competitive ratio with quadratic or linear switching cost is that before the action $\x_t$ is chosen the full function $f_t$ is completely revealed, referred to as the {\it full information setting} in the rest of the paper. 
Recall that the information structure in the OCO problem is fundamentally different; action $\x_t$ at time $t$ is  chosen without any information about $f_t$, and once the action is chosen, one or more gradients $\nabla f_t(.)$ are revealed. We will refer to this as the limited information setting hereafter.
{For minimizing the dynamic regret \eqref{probR:intro1} for OCO-S, a \textit{delayed information} setting has also been considered in \cite{li2020online}, where at time \(t\), an online algorithm \(\mathcal{A}\) takes action \(\x_{t}\) with complete information about  \(f_{1}, \dots, f_{t - 1}\)}. However, a limitation of \cite{li2020online} is that it completely ignores the switching cost of the \(\opt\) and trivially bounds the switching cost of its proposed algorithm. This leads to a large gap between the upper and lower bound on the dynamic regret with respect to \(L, \mu\) for certain problem instances. 

{The full information setting} (as well as the delayed information setting to some extent) is quite imposing for many applications 
and difficult to realize in practice. Thus, in this paper, we study the  OCO-S problem under the limited information setting, similar to the OCO problem. In particular, the limited information setting we consider is as follows: At time \(t\), \(\mathcal{A}\) takes action \(\x_{t}\) with only gradient information about \(f_{t - 1}\) at multiple points. {Moreover, given that the limited information setting is far 
less imposing than the widely considered full information setting, our proposed algorithms are readily applicable to practical settings relevant to 
the performance modelling of computer systems and communication networks, 
where obtaining large amount of sequential information may not be viable.} 


Moving from the full information setting to the limited information setting to solve \eqref{prob:intro1} is challenging on account of the following reasons.
Most algorithms to solve \eqref{prob:intro1} are of the following flavour: given the previous action $\bx_{t-1}$ for algorithm $\cA$, choose the new action $\bx_t$ such that $
f_t(\bx_t) = \beta c(\x_{t}, \x_{t - 1})$, for some constant $\beta > 0$. This algorithm critically needs to know $f_t$ at all points before choosing $\bx_t$. Moreover the analysis using a potential function argument proceeds by comparing the {hitting cost} \(f_{t}(\x_{t})\) of $\cA$ with that of the $\opt$, and that is sufficient since the hitting cost and the switching cost for \(\mathcal{A}\) are multiples of each other.  An alternative approach to construct an algorithm in the full information setting is to simply use  \(\x_{t} = \x_{t}^\star\) \cite{zhang2021revisiting} and use certain structural properties of strongly convex functions.

With the limited information setting, neither can one directly connect the hitting and the switching costs for an online algorithm \(\cA\), since $\cA$ is never aware of $f_t$ before making the decision at time $t$, nor use $\x_{t}^\star$ to make decisions at time $t$. This is the primary challenge we overcome in this paper, where we depart from the potential function approach and proceed as follows:
We propose a simple online gradient descent algorithm that at each time slot makes $\cO(L/\mu)$ number of gradient descent steps in the limited information setting, and upper bound its cost \eqref{cost:intro1}. Next, we separately lower bound the cost \eqref{cost:intro1} of the $\opt$ for any problem instance. Taking the ratio of the two bounds we get our result. As far as we know, this approach has not been used for solving \eqref{prob:intro1}. A byproduct of our approach is that we also get an upper bound on the dynamic regret \eqref{probR:intro1} which is also shown to be order-wise (with respect to the path length $\cP_{T}(\bx^\star)$) optimal when the switching cost is quadratic. 
Similar approach also works for the linear switching cost case as well.

\textbf{Notation:} We use regular font letters to denote scalars, bold-faced lower case letters to denote vectors, and bold-faced upper case letters to denote matrices. The $\ell_2$ norm  of $\x$ is denoted by $\norm{\x}$ and \(\ip{\u}{\v}\) denotes the standard inner product of \(\u\) and \(\v\), defined as \(\ip{\u}{\v} := \u ^ {\T}\v\). Further, $\mathbf{0}$ denotes the vector/matrix of all zeros whose size can be inferred from the context. For a matrix $\X$, $\lambda_{i}(\X)$ denotes its $i^{th}$  eigenvalue. We use the notation \(\X \succeq \mathbf{0}\) \((\X \succ \mathbf{0})\) when \(\X\) is a positive semidefinite (positive definite) matrix which is equivalent to $\lambda_{i}(\X) \ge 0$ $(\lambda_{i}(\X) > 0),$ for all $i$.


\section{Contributions}
\subsection{Quadratic Switching Cost}
Our contributions for solving the OCO-SQ problem, when all $f_t$'s are $\mu$-strongly convex and $L$-smooth, in the limited information setting are as follows.
\begin{enumerate}
	\item 
	We consider an online multiple gradient descent (OMGD) {algorithm} that takes $K$ gradient descent steps (if it can) at  time slot $t$ using the gradients $\nabla f_{t-1}$ at $K$ distinct points with step size equal to $\frac{1}{L}$.
	
	OMGD algorithm essentially tries to choose action $\x_t$ that is  close to the optimizer of the function $f_{t-1}$. It is well-known that for strongly convex functions, one iteration of the usual gradient descent algorithm with appropriate step size has the contraction property that $||\x_i - \x^\star|| \le c ||\x_{i-1} - \x^\star||$ for some $c<1$, where $\x^\star$ is the optimal point and $\{\x_i\}$ are the iterates.  Given that at time $t$, OMGD is trying to get close to the optimizer $\x_{t-1}^\star$ of $f_{t-1}$, using the contraction property and the triangle inequality it is possible to bound the sum of squared distances of the action \(\x_{t}\) produced by OMGD and the  optimizer $\x_{t}^\star$ of $f_{t}$, i.e., the quantity \({\sum_{t = 1}^{T}\norm{\x_{t} - \x_{t}^\star}}^2\) in terms of the squared path length \(\cP_{2, T}^\star = {\sum_{t = 2}^{T}\norm{\x_{t}^\star - \x_{t - 1}^\star}}^2\), a fundamental quantity for OCO. Finally using the $L$-smoothness property, this can be translated to bound the sub-optimality gap $f_t(\x_t) - f_t(\x_t^\star)$. This is the usual recipe to bound the cost of OCO with a gradient descent based algorithm in the limited information setting. 
	
	With the additional switching cost present in \eqref{cost:intro1}, analysing OMGD is lot more challenging, since by its very definition  OMGD disregards the presence of switching cost. It turns out that the contraction property can be leveraged for bounding the switching cost as well. In the analysis, we show that choosing 
	$K = \ceil{\frac{L + \mu}{2\mu}\ln 4}$, the competitive ratio of OMGD algorithm is at most 
	\begin{equation}\label{introcrbound}
		4(L + 5) + \frac{16(L + 5)}{\mu}.
	\end{equation}
	
	To derive this competitive ratio bound, we separately upper bound the cost of OMGD and lower bound the cost of the $\opt$. Lower bounding the cost of the $\opt$ is a non-trivial exercise and as far as we know has not been accomplished before. We also show that a similar competitive ratio guarantee holds for OMGD even against an arbitrary comparator sequence in Remark \ref{rem:comparator}.
	
	\item By constructing  a sequence of $f_t$'s, we show that the competitive ratio of OMGD is at least $1 + \frac{L + 1}{4\mu} + \frac{L + 1}{8}$, showing that order-wise (with respect to \(L, \mu\)), the competitive ratio \eqref{introcrbound} for OMGD cannot be improved. The analysis to derive the lower bound on the competitive ratio for the OMGD algorithm also implies that any online algorithm that at time $t$ tries to `chase' \footnote{Defined after Lemma \ref{lb:omgd}.} the optimizer $\x^\star_{t-1}$ also has the same competitive ratio lower bound as the OMGD algorithm.
	
	\item We also show that the competitive ratio of any  online algorithm is at least $\max\{\Omega\left(L\right), \Omega\big(\frac{L}{\sqrt{\mu}}\big)\}$. These bounds follow rather easily by porting prior results from \cite{goel2019beyond} on the full information model to the limited information model. We in fact conjecture that the competitive ratio of any online algorithm in the limited information setting is at least \(\Omega(\frac{L}{{\mu}})\), which will show a fundamental difference in the dependence of \(\mu\) on the order-wise optimal competitive ratio in the full information setting versus the limited information setting.
	
	\item Our results to bound the competitive ratio of OMGD directly also provide an upper bound on its dynamic regret \eqref{probR:intro1} which is shown to be order-wise optimal (with respect to the path length $\cP_{T}(\bx^\star)$) when the feasible set has a finite diameter. Compared to \cite{li2020online}, our upper bound on the dynamic regret is order-wise tighter in the regime when $L \rightarrow 0$ and $\frac{L}{\mu}$ is finite, and matches in all other regimes.

\end{enumerate}

\subsection{Linear Switching Cost}
Our contributions for solving the OCO-SL problem, when all $f_t$'s are $\mu$-strongly convex and $L$-smooth, in the limited information setting are as follows.
\begin{enumerate}
	\item Similar to the quadratic switching cost case, we consider the OMGD algorithm with \(K = \ceil{\frac{L + \mu}{2\mu} \ln 4}\), and derive an upper bound on its cost. To lower bound the cost of  \(\opt\), we reduce any instance of the linear switching cost problem  to a quadratic switching cost problem, and use the lower bound derived for the quadratic switching cost case.  
	As a result, we get the following upper bound on the  competitive ratio of OMGD  \begin{align*}
		L(\cP_{T}^\star + \norm{\x_{1}^\star}) + \frac{3}{2} \frac{\big(\cP_{T}^\star + \norm{\x_{1}^\star}\big) ^ 2}{\cP_{2, T}^\star + \norm{\x_{1}^\star} ^ 2} + \frac{8L}{\mu} + \frac{12}{\mu} \frac{\cP_{T}^\star + \norm{\x_{1}^\star}}{\cP_{2, T}^\star + \norm{\x_{1}^\star} ^ 2},
	\end{align*}
	where we use the shorthand \(\cP_{T} ^ {\star}\) for \(\cP_{T}(\x^\star)\) throughout, and \(\cP_{2, T} ^ \star = \sum_{t = 2} ^ T \norm{\x_{t}^\star - \x_{t - 1}^\star} ^ 2\). Different from the quadratic switching cost problem, where the competitive ratio depends only on \(L, \mu\), the competitive ratio for the linear switching cost problem depends on problem specific quantities \(\cP_{T}^\star\) and \(\cP_{2, T} ^ \star\), in addition to \(L\), \(\mu\). 
	\item By constructing a sequence of \(f_{t}\)'s that are \(\mu\)-strongly convex and \(\mu\)-smooth, we show that the competitive ratio of any online algorithm can be lower bounded by \begin{align*}
		\Omega \bigg( \mu(\cP_{T}^\star + \norm{\x_{1}^\star}) + \frac{3}{2} \frac{\big(\cP_{T}^\star + \norm{\x_{1}^\star}\big) ^ 2}{\cP_{2, T}^\star + \norm{\x_{1}^\star} ^ 2} + \frac{12}{\mu} \frac{\cP_{T}^\star + \norm{\x_{1}^\star}}{\cP_{2, T}^\star + \norm{\x_{1}^\star} ^ 2}\bigg),
	\end{align*}
	which order-wise (with respect to \(\cP_{T}^\star\) and \(\cP_{2, T}^\star\)) matches with the upper bound on the competitive ratio of OMGD.

	Recall that with the full information setting, algorithm with competitive ratio of $2$ \cite{bansal20152} is known even when the functions $f_t$'s are just convex and $d=1$. Thus, our result shows that the limited information setting is fundamentally different than the full information setting when the switching cost is linear, and the competitive ratio of any online algorithm is a function of the path length  \(\cP_{T}^\star\) or the squared path length \(\cP_{2, T}^\star\).
	
\end{enumerate}
\subsection{ Comparison with prior work}

\subsubsection{\textup{Quadratic Switching Cost}\\}
For $\mu$-strongly convex functions $f_t$'s, in the full information setting, the optimal competitive ratio for \eqref{prob:intro1} is $\Theta (\frac{1}{\sqrt{\mu}})$ as \(\mu \to 0\) \cite{goel2019beyond}. Our results show that the limited information setting is fundamentally different and any online algorithm has to have a competitive ratio growing linearly in $L$ and $\frac{L}{\sqrt{\mu}}$. 
Surprisingly, when $L= c \mu$, {for some finite constant \(c \geq 1\)} (that is independent of \(T\) and the problem parameters, i.e., \(L, \mu\), etc.), the competitive ratio of the OMGD algorithm is $\mathcal{O}(\frac{1}{\mu})$ as \(\mu \to 0\) similar to the OBD algorithm \cite{chen2018, goel2019online} that requires full information about the functions $f_t$'s, which is rather remarkable. 

For the delayed information setting, \cite{li2020online} considered the problem of minimizing dynamic regret \eqref{probR:intro1} over a compact feasible set. The online gradient descent algorithm that takes only one gradient step achieves a regret of $\mathcal{O}\big(\frac{G}{1 - \sqrt{1 - \mu / L}} \big(1+\frac{1}{L}\big)(\mathcal{P}_T^\star + D)\big)$, while the lower bound on the regret for \(\mu\)-strongly convex and \(\mu\)-smooth \(f_t\)'s is $\Omega\left(\zeta D(\norm{\x_{1}^\star} + \mathcal{P}_{T}^\star)\right)$, where  \(\zeta = \frac{\mu ^ {3}(1 - \rho)^{2}}{32(\mu + 1)^{2}}, \rho = \frac{\sqrt{\mu + 4} - \sqrt{\mu}}{\sqrt{\mu + 4} + \sqrt{\mu}}\), $D$ is the diameter of the feasible set $\cX$, \(\max_{\x \in \mathcal{X}}\norm{\nabla f_{t}(\x)} \le G\), and $\mathcal{P}_{T}^{*} = \sum_{t = 2}^{T} \norm{\x_{t}^\star - \x_{t - 1}^\star}$. Thus, when $L\rightarrow 0$ and \(\frac{L}{\mu}\) is finite, the upper and lower bounds of \cite{li2020online} are not close. This gap is primarily the result of \cite{li2020online}'s analysis that ignores the switching cost of the \(\opt\) and bounds the switching cost of its proposed algorithm trivially. In comparison to \cite{li2020online}, we show by a tighter analysis that the OMGD algorithm achieves a dynamic regret which is order-wise (in terms of \(\cP_{T}^\star\)) the same as the lower bound of \cite{li2020online} and has a superior dependence of \(L, \mu\) than the factor \(\frac{G}{1 - \sqrt{1 - \mu / L}} \big(1+\frac{1}{L}\big)\) of \cite{li2020online}, even in the limited information setting.

In comparison to the most related prior work \cite{chen2018, goel2019beyond, zhang2021revisiting} on the competitive ratio analysis of OCO-SQ,  our analysis strategy for deriving an upper bound on the competitive ratio \eqref{prob:intro1} for the OMGD algorithm  is {novel}. We separately upper bound the cost of the OMGD algorithm and lower bound the cost of $\opt$, while the known analysis methods use a potential function argument \cite{chen2018, goel2019beyond} or compare the relative costs of the algorithm against the \(\opt\)  using structural properties of strongly convex functions \cite{zhang2021revisiting}. Lower bounding the cost of $\opt$ for the quadratic switching cost is non-trivial while taking into account the quadratic switching cost and as far as we know has not been accomplished before.

\subsection{Linear Switching Cost}
In the full information setting, the competitive ratio of any online algorithm for solving \eqref{prob:intro1} has a lower bound \(\cO(\big(\frac{L}{\mu}\big) ^ {\frac{1}{3}}\big)\) while the COBD algorithm \cite{argue2020dimension} attains a competitive ratio \(\cO(\big(\frac{L}{\mu}\big) ^ {\frac{1}{2}})\). However, in the limited information setting our results show that it is not possible to attain dimension-free bounds on the competitive ratio and any online algorithm has a competitive ratio that explicitly depends on \(\cP_{T}^\star\) and \(\cP_{2, T}^\star\). Moreover, in comparison to the most related prior work \cite{argue2020dimension, bansal2015} on the linear switching cost problem, our analysis strategy is {different}. Motivated by our results on the quadratic switching cost problem, we upper bound the cost of the OMGD algorithm and lower bound the cost of the \(\opt\) for any given problem instance by reducing the linear switching cost problem to an instance of the quadratic switching cost problem.

{Our contributions on the competitive ratio of an algorithm for the quadratic and the linear switching cost problems in the limited information setting, and our comparison against prior works are summarized in Table \ref{tab:comparison_prior_works_rebuttal}. As mentioned in the introduction, when the functions \(f_{t}\)'s are convex, \cite{chen2018smoothed} showed that the lower bound on the competitive ratio is \(\Omega(\sqrt{d})\), therefore no online algorithm \(\cA\) can attain a constant competitive ratio that is independent of \(d\). Therefore, we exclude the case when \(f_{t}\)'s are convex and \(d\) is arbitrary from the table. Further, \cite{zhang2021revisiting, goel2019beyond} considered $f_{t}$'s to satisfy the \(\mu\)-quadratic growth condition, i.e. $f_{t}(\x) - f_{t}(\x_{t}^\star) \ge \frac{\mu}{2}\norm{\x - \x_{t}^\star} ^ {2}$ for all \(\x \in \cX\), in addition to convexity \cite{zhang2021revisiting}, quasiconvexity \cite{goel2019beyond}. In our context, for simplicity, in Table \ref{tab:comparison_prior_works_rebuttal} we state the result for strongly convex functions as they satisfy the quadratic growth condition.}
\begin{tiny}
	\begin{table}[!htb]
		\centering
		\begin{tabular}{|p{1.7cm}|p{1.8cm}|p{1.8cm}|p{1.5cm}|p{3cm}|p{3cm}|}
			\hline
			information setting & assumption on \(f_{t}\) & dimension (\(d\)) &  switching cost & upper bound on \(\mu_{\mathcal{A}}\) & lower bound  on \(\mu_{\mathcal{A}}\) \\
			\hline
			full & convex & 1 & linear & 2 \cite{bansal2015} & 2 \cite{antoniadis2017tight} \\ \hline
			full & \(\alpha\)-polyhedral\(^{*}\) & arbitrary & linear & \(3 + \frac{8}{\alpha}\) \cite{chen2018smoothed} & not given \\ \hline 
			full & \(\alpha\)-polyhedral\(^{*}\) & arbitrary & linear & \(\max(1, \frac{2}{\alpha})\) \cite{zhang2021revisiting} & not given \\ \hline
			full & \(\mu\)-strongly convex + \(L\)-smooth & arbitrary & linear & \(\cO({\frac{L}{\mu}}) ^ {\frac{1}{2}}\) \cite{argue2020dimension} & \(\Omega(\frac{L}{\mu}) ^ {\frac{1}{3}}\) \cite{argue2020dimension} \\
			\hline
			limited (this work) & \(\mu\)-strongly convex + \(L\)-smooth & arbitrary & linear & depends on \(\mathcal{P}_{T}^\star\), \(\mathcal{P}_{2, T}^\star\) (Theorem \ref{thm:cromgd_linear}) & depends on \(\mathcal{P}_{T}^\star\), \(\mathcal{P}_{2, T}^\star\) (Lemma \ref{lem:lb_linear}) \\ \hline
			full & \(\mu\)-strongly convex & arbitrary & quadratic & \(3 + \mathcal{O}(\frac{1}{\mu})\) \cite{goel2019online} & \(\Omega(\frac{1}{\sqrt{\mu}})\) as \(\mu \to 0\)\cite{goel2019beyond} \\
			\hline
			full & \(\mu\)-strongly convex & arbitrary &  quadratic & \(1 + \frac{4}{\mu}\) \cite{zhang2021revisiting} & \(\Omega(\frac{1}{\sqrt{\mu}})\) as \(\mu \to 0\)\cite{goel2019beyond}  \\
			\hline
			full & \(\mu\)-strongly convex & arbitrary & quadratic & \(\mathcal{O}(\frac{1}{\sqrt{\mu}})\) as \(\mu \to 0\) \cite{goel2019beyond} & \(\Omega(\frac{1}{\sqrt{\mu}})\)  as \(\mu \to 0\)\cite{goel2019beyond}\\
			\hline
			limited (this work) & \(\mu\)-strongly convex + \(L\)-smooth & arbitrary & quadratic & \(\mathcal{O}(\frac{L}{\mu})\) as \(\mu \to 0\) (Theorem \ref{thm:cromgd}) & \(\Omega(\frac{L}{\sqrt{\mu}})\) as \(\mu \to 0\) (Lemma \ref{modified_preliminary_lower_bound}) \\ \hline
		\end{tabular}
		\caption{Comparison of the competitive ratio of the proposed algorithm with existing works.}
		\label{tab:comparison_prior_works_rebuttal}
	\end{table}
\end{tiny}
\def\thefootnote{*}\footnotetext{{A function \(f: \mathcal{X} \to \Rn\) is \(\alpha\)-polyhedral if \(f(\x) - f(\x^\star) \ge \alpha \norm{\x - \x ^ \star}\) for all \(x \in \cX\), where \(\x^\star := \argmin_{\x \in \cX} f(\x)\).}}



%

	\section{Problem Setting}
	{\bf OCO with Switching Cost (OCO-S) Problem:} Consider that at time $t=1, 2, \dots, T$, a convex function $f_t:\bbR^d \rightarrow \bbR^+$ is going to be revealed and an online algorithm $\cA$'s objective is to choose action $\bx_t \in \cX \subseteq \bbR^d$ so as to the minimize the overall cost 
	\begin{equation}\label{cost:intro}
		C_\cA = \sum_{t=1}^T  f_t(\bx_t) + c(\x_{t}, \x_{t - 1}),
	\end{equation} 
	which comprises of the \textbf{hitting cost} \(f_{t}(\x_{t})\) and a non-negative \textbf{switching cost} \(c(\x_{t}, \x_{t - 1})\).
	Compared to OCO, there is an additional penalty in \eqref{cost:intro} if $\cA$ changes its decisions between time slots $t-1$ and $t$. 
	The goal for $\cA$ is to  either minimize the 
	dynamic regret 
	\begin{equation}\label{probR:intro}
		R_d(\cA) = \max_{f_t}(C_\cA - C_\opt),
	\end{equation}
	or the competitive ratio
	\begin{equation}\label{prob:intro}
		\text{cr}_\cA = \max_{f_t}\frac{C_\cA}{C_\opt},
	\end{equation}
	{where \(C_\opt\) denotes the cost of the $\opt$, that knows all \(f_{t}\)'s ahead of time,  for solving \eqref{cost:intro}}. 
	In this paper, we consider two forms of the switching cost (a) quadratic switching cost, where \(c(\x_{t}, \x_{t - 1}) = \frac{1}{2}\norm{\x_{t} - \x_{t - 1}} ^ 2\), and (b) linear switching cost, where \(c(\x_{t}, \x_{t - 1}) = \norm{\x_{t} - \x_{t - 1}}\). For (a) and (b) both, we primarily focus on finding online algorithms with small competitive ratios. Moreover, for (a), we also show how to bound the dynamic regret using the corresponding analysis for the competitive ratio.

	Throughout the paper we shall assume that \(\mathcal{X}\) is a non-empty closed convex set and contains the point \(\mathbf{0}\). An additional assumption about the compactness of \(\mathcal{X}\) is made in Sec. \ref{sec:regret} while bounding the dynamic regret. 
	Moreover, for bounding the cost attained by our proposed algorithm $\cA_o$, \(C_{\cA_o}\) \eqref{cost:intro}, we assume that the optimizers \(\x_{t}^\star \in \text{int}(\mathcal{X})\). Discussion about this assumption can be found in Remark \ref{rem:int}.
	Note that all algorithms for \eqref{cost:intro}, start from the same starting point \(\x_{0} \in \mathcal{X}\).

	
	
	
	For solving \eqref{prob:intro}, as far as we know only the {\it full information setting} is assumed, where $f_t$ is revealed completely {\bf before} $\bx_t$ is chosen. Because of the switching cost, even in the full information setting, solving \eqref{prob:intro} remains non-trivial.
	The full information setting requires incredibly large amount of information compared to the limited information setting in which OCO is studied, and is not reasonable to assume for many applications. Thus, in this paper we intend to solve \eqref{prob:intro} under the limited information setting, where in particular we make the following assumption.
	
	\begin{assumption}\label{ass:info}
		At time \(t\), \(\cA\) takes action \(\x_{t}\) with only gradient information about \(f_{t - 1}\) at multiple points. 
	\end{assumption} 

	From \cite{goel2019beyond}, we know that if functions $f_t$'s are just convex and the switching cost is quadratic, the competitive ratio \eqref{prob:intro}  of any online algorithm even in the full information setting is unbounded. Similarly, from \cite{chen2018smoothed}, we know that the competitive ratio of any online algorithm when the switching cost is linear in the full information setting is \(\Omega(\sqrt{d})\) and is therefore arbitarily large for high problem dimensionality. Thus, similar to prior work  for solving \eqref{prob:intro}, we assume the following:
	
	\begin{assumption}\label{function_structure}
		For all \(1 \leq t \leq T\), the function \(f_{t}: \mathcal{X} \to \mathbb{R}^{+}\) is \(L\)-smooth and \(\mu\)-strongly convex. Mathematically, this implies \(
		\frac{\mu}{2} \norm{\y - \x} ^ {2} \leq f_t(\y) - f_t(\x) - \ip{\nabla f_t(\x)}{\y - \x} \leq \frac{L}{2} \norm{\y - \x}^2,
		\)
		for all \(\x, \y \in \mathcal{X}\).
	\end{assumption}
	Note that with the full information setting, $L$-smoothness is not necessary, however, we need it since we are working with the limited information setting.
	
	We next propose a simple algorithm and bound its competitive ratio for both the quadratic switching cost as well as linear switching cost separately. The challenge in proposing algorithms and analysing them in the limited information setting compared to the full information setting has been discussed in the Introduction \ref{sec:introduction}.
	
	\section{Algorithm}
	Let $\px{\cX}{\bx} := \argmin_{\y \in \cX} \norm{\y - \x}$ be the projection of point $\bx$ onto set $\cX$.
	We consider the online multiple gradient descent (OMGD) algorithm  (Algorithm \ref{alg:omgd}) for minimizing the cost \eqref{cost:intro} that was originally proposed for minimizing the dynamic regret for OCO. The OMGD algorithm is a very simple algorithm that takes $K$ gradient descent steps (if it can) with respect to the objective function $f_{t-1}$ at each time $t$. Notably, it is disregarding the presence of the switching cost in \eqref{cost:intro}. The presence of the switching cost with OCO-S makes the analysis of OMGD different from \cite{zhang2017improved}.
	
	\begin{algorithm}
		\caption {Online Multiple Gradient Descent (OMGD) for OCO-S}
		\begin{algorithmic}[1]
			\STATE\textbf{Input:} Problem parameters, i.e., $\x_{0} (\text{starting point)}, \mathcal{X} (\text{feasible set}), L(\text{smoothness parameter})$;
			\STATE Choose \(\x_{1} = \x_0\);
			\STATE\textbf{for} \(t = 2, \ldots, T\)
			\STATE\hspace{3mm} Set \(\z_{t} ^ {(0)} = \x_{t - 1}\);
			\STATE\hspace{3mm} \textbf{for} \(k = 1, \ldots, K\)
			\STATE\hspace{3mm}\hspace{3mm} Query \(\nabla f_{t - 1}(\z_{t} ^ {(k - 1)})\);
			\STATE\hspace{3mm}\hspace{3mm} Update \(\z_{t} ^ {(k)} = \px{\mathcal{X}}{\z_{t} ^ {(k - 1)} - \frac{1}{L}\nabla f_{t - 1}(\z_{t} ^ {(k - 1)})}\);
			\STATE\hspace{3mm}\textbf{end}
			\STATE\hspace{3mm} Choose \(\x_{t} = \z_{t} ^ {(K)}\);
			\STATE\textbf{end}
		\end{algorithmic}
		\label{alg:omgd}
	\end{algorithm}
	
	\begin{rem} Eventually, we are going to select \(K =  \ceil{\frac{L + \mu}{2\mu}\ln 4}\)  for the OMGD algorithm thus requiring the knowledge of both $L,\mu$. Note that for any gradient descent algorithm, the step size is typically a function of $L$. Thus, over and above the usual assumption of the knowledge of $L$, we are also assuming that $\mu$ is known ahead of time.
	\end{rem} 
	\begin{rem}\label{rem:int}
		{We discuss the reason to assume that the optimizers \(\x_{t} ^\star \in \text{int}(\mathcal{X})\). We require this assumption in Theorem \ref{thm:cromgd} in order to set \(\nabla f_{t}(\x_{t}^\star) = \mathbf{0}\), and get an upper bound on the cost \({C}_{\mathcal{A}_{o}}\)  \eqref{cost:intro} (where \(\mathcal{A}_o\) is the OMGD algorithm) that is independent on the norm of the gradient \(\nabla f_{t}(\x_{t}^\star)\). If  \(\x_{t} ^\star \notin \text{int}(\mathcal{X})\), we show in \ref{app:remark_6} that as long as the quantity 
			\(\sum_{t = 1} ^ {T} \norm{\nabla f_{t}(\x_{t}^\star)}^{2}\) is sufficiently small, i.e., \(\sum_{t = 1}^{T} \norm{\nabla f_{t}(\x_{t}^\star)}^{2} = \mathcal{O}\big(\cP_{2, T}^\star)\), we can get the same order-wise (with respect to \(L, \mu\)) bound as Theorem \ref{thm:cromgd}. 
			It is worth noting that for deriving a lower bound in Lemma \ref{lem:lbOPT} on the cost \eqref{cost:intro} incurred by \(\opt\), we do not make the assumption  that \(\x_{t}^\star \in \text{int}(\mathcal{X})\).}
\end{rem}
\section{Quadratic Switching Cost}
In this section, we consider \eqref{cost:intro} in the context of the quadratic switching cost, i.e., the cost of an online algorithm \(\cA\) is given by \begin{align}\label{eq:cost_quadratic}
	C_{\cA} = \sum_{t = 1} ^ T f_{t}(\x_{t}) + \frac{1}{2}\norm{\x_{t} - \x_{t - 1}} ^ 2.
\end{align}
{Note that in \eqref{eq:cost_quadratic}, we can also include a scaling factor $\delta>0$ for the switching cost $ \norm{\x_{t} - \x_{t - 1}}^{2}$ that can tradeoff the relative importance of the two costs. However, that is not necessary since equivalently, $f_t$'s can be scaled by $\frac{1}{\delta}$, changing their strong convexity parameter appropriately. Since our guarantees will be a function of the strong convexity parameter, we just consider \eqref{eq:cost_quadratic}.}

\subsection{Competitive Ratio}

\subsubsection{{\textup{Upper Bounding the Cost of OMGD\nopunct}}\\}

Let $\cA_o$ be the OMGD algorithm. 
We now proceed towards upper bounding the total cost \eqref{eq:cost_quadratic} of $\cA_o$. Let, the action chosen by $\cA_o$ at time $t$ be $\bx_t$. The following Lemma establishes a bound on the quantity \(\sum_{t = 1} ^ {T} \norm{\x_{t} - \bx_t^\star} ^ {2}\), where \(\x_{t}^\star = \argmin_{\x \in \mathcal{X}} f_{t}(\x)\), which can be interpreted as the cumulative progress of $\cA_o$, measured as squared distance from the optimizers \(\{\bx_t^\star\}_{t =1} ^ {T}\). This quantity will be useful in bounding both the total hitting cost and the total switching cost of $\cA_o$.
\begin{lemma}\label{squared_distance_lemma}
	\textit{Under assumption \ref{function_structure}, for $\cA_o$ (Algorithm \ref{alg:omgd}) with 
		\(K = \ceil {\frac{L + \mu}{2\mu}\ln 4}\), we have \begin{align}\label{actual_squared_bound}
			\sum_{t = 1} ^ {T} \norm{\x_{t} - \bx_t^\star} ^ {2} \leq 2 \norm{\x_{1} - \bx^\star_{1}} ^ 2 + 4 \P_{2, T} ^\star,
		\end{align}
		where 		
		$\P_{2, T} ^\star = \sum_{t=2}^T \norm{\x_{t}^\star - \bx_{t-1}^\star} ^ {2}$.
	}
\end{lemma}
$\P_{2, T} ^\star$ is the squared {path length} that captures the squared distance between consecutive optimizers and is a quantity that 
naturally controls the cost of any algorithm. 
\begin{proof}
	At time \(t \geq 2\), upon taking a gradient descent step from \(\z_{t} ^ {(k - 1)}\) to \(\z_{t} ^ {(k)}\), the following contraction holds for $\cA_o$ from Lemma \ref{lem:contraction} (refer \ref{app:lem:contraction}) with \(\eta = \frac{1}{L}\): \[
	||{\z_{t} ^ {(k)} - \bx^\star_{t - 1}}|| ^ {2} \leq \bigg(\frac{L - \mu}{L + \mu}\bigg) ||{\z_{t} ^ {(k - 1)} - \bx^\star_{t - 1}} ||^ 2.
	\]
	Therefore, the net contraction after the end of \(k\) gradient steps at time \(t\) is  \[
	||{\z_{t} ^ {(k)} - \bx^\star_{t - 1}}|| ^ {2} \leq \bigg(\frac{L - \mu}{L + \mu}\bigg) ^ {k} ||{\z_{t} ^ {(0)} - \bx^\star_{t - 1}}|| ^ 2.
	\]
	Since, \(\z_{t} ^ {(0)} = \x_{t - 1},  \x_{t} = \z_{t} ^ {(K)}\), the contraction for \(\x_t\) compared to \(\x_{t - 1}\) is given by \(
	||{\x_t - \bx^\star_{t - 1}}|| ^ 2 \leq \big(\frac{L - \mu}{L + \mu}\big) ^ {K} ||{\x_{t - 1} - \bx^\star_{t - 1}}|| ^ 2.
	\)
	
	Since \(e^{x} \geq 1 + x,\) for all \(x \in \mathbb{R}\), choosing \(K =  \ceil{\frac{L + \mu}{2\mu}\ln 4}\), we get \begin{align}\nn
		\bigg(1 - \frac{2\mu}{L + \mu}\bigg) ^ {K} \leq \exp \bigg(-\frac{2\mu}{L + \mu} \left \lceil{\frac{L + \mu}{2\mu}\ln 4} \right \rceil \bigg) \leq \exp \bigg(-\frac{2\mu}{L + \mu}{\frac{L + \mu}{2\mu}\ln 4}\bigg) = \frac{1}{4}.
	\end{align}
	Therefore, we get 
	\vspace{-0.1in}\begin{align}\label{one_fourth_0}
		||{\x_{t} - \bx^\star_{t - 1}}|| ^ {2} \leq \frac{1}{4} ||{\x_{t - 1} - \bx^\star_{t - 1}}|| ^ 2.
	\end{align}
	It is important for further exposition that the constant in the contraction \eqref{one_fourth_0} is a constant. If $K=1$ in the OMGD algorithm, then the  constant will depend on $\mu$ and $L$ and will limit further analysis. 
	Using \eqref{one_fourth_0}, we have the following: \begin{align}
		\sum_{t = 2} ^ {T} \norm{\x_{t} - \bx_t^\star} ^ {2} &\stackrel{(a)}\leq \sum_{t = 2} ^ {T} (2 \norm{\x_{t} - \bx^\star_{t - 1}} ^ 2 + 2 \norm{\bx_t^\star - \bx^\star_{t - 1}} ^ 2)\nn,\\
		&\leqtext{\eqref{one_fourth_0}} \frac{1}{2} \sum_{t = 2} ^ {T} \norm{\x_{t - 1} - \bx^\star_{t - 1}} ^ {2} + 2 \P_{2, T} ^\star \leq \frac{1}{2} \sum_{t = 1} ^ {T} \norm{\x_{t} - \bx_t^\star} ^ {2} + 2\P_{2, T} ^\star, \label{rearranging_}\end{align}
	where to get  $(a)$  we have used the Triangle inequality followed by \(\norm{\u + \v} ^ {2} \leq 2(\norm{\u} ^ {2} + \norm{\v} ^ {2})\).
	
	Rearranging \eqref{rearranging_}, we obtain \(
	\sum_{t = 2} ^ {T} ||{\x_{t} - \bx_t^\star}|| ^ 2 \leq ||{\x_{1} - \bx^\star_{1}||} ^ 2 + 4 \P_{2, T} ^\star 
	\implies\sum_{t = 1} ^ {T} ||{\x_{t} - \bx_t^\star}|| ^ {2} \leq 2 ||{\x_{1} - \bx^\star_{1}}|| ^ 2 + 4 \P_{2, T} ^\star.
	\)
	This completes the proof.
\end{proof}
\begin{rem}\label{rem:NAG}
	From the discussion above, we know that at time \(t\), OMGD can obtain \(\frac{1}{4}\)-th contraction by performing gradient descent, using at most \(K = \cO(\frac{L}{\mu})\) gradient steps. However, by using Nesterov Accelerated Gradient (NAG), i.e., at time \(t\), instead of performing \(K\) gradient steps, execute \(K\) steps of NAG, we can show that the number of steps can be substantially reduced. In \ref{app:NAG}, we show that \(K = {\cO}(\sqrt{\frac{L}{\mu}}\log {\frac{L}{\mu}})\) gradient steps are sufficient to obtain \(\frac{1}{4}\)-th contraction \eqref{one_fourth_0}.
\end{rem}
The next Lemma bounds the total hitting cost \(\sum_{t = 1} ^ {T} f_{t}(\x_{t})\) of $\cA_o$.
\begin{lemma}\label{lemma_hitting_cost}
	\textit{Under assumption \ref{function_structure}, the total hitting cost \(\sum_{t = 1} ^ {T} f_{t}(\x_{t})\) of $\cA_o$  with \(K = \ceil{\frac{L + \mu}{2\mu}\ln 4}\) is bounded by 
		$$
		\sum_{t = 1} ^ {T} f_{t}(\x_{t}) \leq \sum_{t = 1} ^ {T} f_{t}(\bx_t^\star) + \frac{1}{2\alpha} \sum_{t = 1} ^ {T} \norm{\nabla f_{t}(\bx_t^\star)} ^ {2} + (L + \alpha)(\norm{\x_{1} - \bx^\star_{1}} ^ {2} + 2\P_{2, T} ^\star),$$
		for any \(\alpha > 0\).}
\end{lemma}

The proof of Lemma \ref{lemma_hitting_cost} follows from the smoothness of  \(f_{t}\)'s and can be found in  \ref{app:hitting_cost}. Note that for the OMGD algorithm, the contraction obtained in \eqref{one_fourth_0} is with respect to the preceding optimizer \(\x_{t - 1}^\star\), and the fact that all functions $f_t$'s have Lipschitz continuous gradients, it is not surprising that we can upper bound the total hitting cost of $\cA_o$. What is not obvious is that it can also be used 
to upper bound the total switching cost of $\cA_o$ which we do in the following manner in Lemma \ref{lemma_switching_cost}: (a) By applying the Triangle inequality and the contraction result \eqref{one_fourth_0}, we relate \(\sum_{t = 2} ^ T \norm{\x_{t} - \x_{t - 1}} ^ 2\) with \(\sum_{t = 1} ^ T \norm{\x_{t} - \x_{t} ^ \star} ^ 2\); (b) Finally, applying the result of Lemma \ref{squared_distance_lemma}, we relate \(\sum_{t = 1} ^ T \norm{\x_{t} - \x_{t}^\star} ^ 2\) with \(\cP_{2, T}^\star\). 

Recall that the OMGD algorithm by itself does not have any feature to control the switching cost.
\begin{lemma}\label{lemma_switching_cost}
	\textit{Under assumption \ref{function_structure}, for $\cA_o$  with \(K = \ceil{\frac{L + \mu}{2\mu}\ln 4}\), the total switching cost is bounded by 
		\[
		\sum_{t = 1} ^ {T} \frac{1}{2}\norm{\x_{t} - \x_{t - 1}} ^ {2} \leq 5 \norm{\x_{1} - \bx^\star_{1}} ^ {2} + 10 \P_{2, T} ^\star.
		\]
	}
\end{lemma}
\begin{proof}
	Using the triangle inequality followed by \(\norm{\u + \v} ^ {2} \leq 2 (\norm{\u}^{2} + \norm{\v}^{2})\), we obtain the following for all \(t \geq 2\):
	\[
	\norm{\x_{t} - \x_{t - 1}} ^ {2}  \leq 2 \norm{\x_{t} - \bx^\star_{t - 1}} ^ {2} + 2 \norm{\x_{t - 1} - \bx^\star_{t - 1}} ^ 2  \leqtext{\eqref{one_fourth_0}} \frac{5}{2} \norm{\x_{t - 1} - \bx^\star_{t - 1}} ^ 2.
	\]
	Hence, the total switching cost of $\cA_o$ can be bounded by \begin{align}\nn
		\sum_{t = 1} ^ {T} \frac{1}{2} \norm{\x_{t} - \x_{t - 1}} ^ {2} &\stackrel{(a)}= \sum_{t = 2} ^ {T} \frac{1}{2}\norm{\x_{t} - \x_{t - 1}} ^ {2}, \\ &\stackrel{(b)}\leq \sum_{t = 1} ^ {T - 1} \frac{5}{2} \norm{\x_{t} - \bx_t^\star} ^ {2} ,\nn \\ &\leqtext{\eqref{actual_squared_bound}} 5 \norm{\x_{1} - \bx^\star_{1}} ^ {2} + 10 \P_{2, T} ^\star,	\nn				
	\end{align}
	where \((a)\) follows since \(\mathcal{A}_o\) chooses \(\x_{1} = \x_{0}\), \((b)\) follows from \(\norm{\x_{t} - \x_{t - 1}}^{2} \leq \frac{5}{2}\norm{\x_{t - 1} - \x_{t - 1}^\star}^{2}\)  established above. This completes the proof.
\end{proof}
Finally, combining Lemma \ref{lemma_hitting_cost} and Lemma \ref{lemma_switching_cost}, we get the following Lemma that bounds the total cost of \(\cA_o\).
\begin{lemma}\label{lem:upOMGD}
	\textit{Under Assumption \ref{function_structure}, the total cost \eqref{eq:cost_quadratic} of $\cA_o$ with \(K = \ceil{\frac{L + \mu}{2\mu}\ln 4}\) is bounded by \[
		C_{\cA_o} \leq \sum_{t = 1} ^ {T} f_{t}(\bx_t^\star) + \frac{1}{2\alpha} \sum_{t = 1} ^ {T} \norm{\nabla f_{t}(\bx_t^\star)} ^ {2} + (L + \alpha + 5)(\norm{\x_{1} - \bx^\star_{1}} ^ 2 + 2\P_{2, T} ^\star  ),\]
		for any \(\alpha > 0\).}
\end{lemma}
\subsubsection{{\textup{Lower bounding the cost of $\opt$\nopunct}}\\}
In this section, we lower bound the cost of $\opt$ \eqref{eq:cost_quadratic}. 
Without loss of generality, the starting location for all algorithms is \(\x_{0} = \mathbf{0}\). For deriving Lemma \ref{lem:lbOPT}, we do not need the assumption that $\bx_t^\star \in \text{int}(\bX)$.
\begin{lemma}\label{lem:lbOPT}
	\textit{Under assumption \ref{function_structure}, the cost incurred by the $\opt$ \eqref{eq:cost_quadratic} satisfies \begin{align}\label{actual_lower_bound}
			{C_\opt \geq \sum_{t = 1} ^ {T} f_t(\bx^\star_t) + \frac{\mu}{2(\mu + 4)} \big(\P_{2, T} ^\star + \norm{\bx^\star_{1}} ^ {2}\big).}
		\end{align}
	}
\end{lemma}
To prove Lemma \ref{lem:lbOPT}, we do the following: Using the strong convexity of $f_t$'s, first, we lower bound $C_\opt$ by an optimization problem of the following quadratic form: $$C_{\opt} \geq \sum_{t = 1} ^ {T} f_{t}(\bx_t^\star) + \min_{\{\tilde{\x}_{k} \in \mathbb{R} ^ {T}\}_{k = 1} ^ {d}}  \sum_{k = 1} ^ {d} \psi(\tilde{\x}_{k}),$$
where we perform a change of variable by \(\tilde{\x}_{k} = [x_{1} ^ {(k)}, \ldots, x_{T} ^ {(k)}] ^ {\T} \in \mathbb{R}^{T}\), where \(T\) is the time horizon. The function \(\psi(\tilde{\x}_{k})\) is defined as \[\psi(\tilde{\x}_{k}) := \sum_{t = 1} ^ {T}\big(\frac{\mu}{2} (x_{t} ^ {(k)} - x_t ^ {\star(k)}) ^ {2} + \frac{1}{2} (x_{t} ^ {(k)} - x_{t - 1} ^ {(k)}) ^ {2}\big) = \frac{\mu}{2} (\tilde{\x}_{k} - \tilde{\bx}_{k}^\star) ^ {\T} (\tilde{\x}_{k} - \tilde{\bx}_{k}^\star) + \frac{1}{2} \tilde{\x}_{k} ^ {\T}\B \tilde{\x}_{k},\] where \(\tilde{\x}_k^\star = [x_{1}^{\star(k)}, \dots, x_{T}^{\star(k)}]\), and the matrix $\B$ given by \eqref{b_def} is 
of dimension \(T \times T\).
Clearly, \(\psi(\tilde{\x}_{k})\) is a quadratic function in \(\tilde{\x}_{k}\). Further, we define a matrix \(\A := \B + \mu \I\) and bound the eigenvalues of \(\A\) using the Gershgorin's Circle Theorem \cite[Thm. 6.1.1]{horn2012matrix} by exploiting the special structure of $\B$. In particular, we show that 
\(\nn\lambda_{i}\big(\mu\A ^ {-1} + \frac{1}{\mu + 4}\A\big) \leq 1 + \frac{\mu}{\mu + 4},\) for all \(1 \le i \le T\). Combining these results, we get the required lower bound \eqref{actual_lower_bound}. The full proof is rather long and can be found in  \ref{app:lbOPT}. \begin{rem}When \(f_{t} = f,\) for all \(1 \leq t \leq T\), a simpler proof is possible for Lemma \ref{lem:lbOPT} and can be found in Lemma \ref{f_tsame} of  \ref{app:lbOPT}.

\begin{rem}[\textit{Upper bound on the cost incurred by the $\opt$}]
	Unlike the lower bound (Lemma \ref{lem:lbOPT}), the upper bound on $C_\opt$ is straightforward. Observe that setting \(\x_{t} = \bx_t^\star\) in \eqref{eq:cost_quadratic}, gives us the following: \(
	C_\opt \leq \sum_{t = 1} ^ {T} \big(f_{t}(\bx_t^\star) + \frac{1}{2} \norm{\bx_t^\star - \bx^\star_{t - 1}} ^ {2}\big) = \sum_{t = 1} ^ {T} f_{t}(\bx_t^\star) + \frac{1}{2}(\P_{2, T} ^\star + \norm{\bx^\star_{1}} ^ {2})
	\), which is similar to \eqref{actual_lower_bound}.
\end{rem}

\end{rem}
\subsubsection{\textup{{Bounding the Competitive Ratio of OMGD\nopunct}}\\} 
Combining Lemma \ref{lem:upOMGD} and Lemma \ref{lem:lbOPT}, we upper bound the competitive ratio  of $\cA_o$ (the OMGD algorithm) as follows.
\begin{theorem}\label{thm:cromgd}
\textit{Under assumption \ref{function_structure}, and assuming that the minimizers \(\bx_t^\star \in \text{int}(\mathcal{X})\), the competitive ratio  of the OMGD algorithm $\cA_o$ satisfies}  \[
{\text{cr}_{\cA_o}=\frac{C_{\cA_o}}{C_\opt} \leq 4(L + 5) + \frac{16(L + 5)}{\mu}.}
\]
\end{theorem}
\begin{proof}
Using the result of Lemma \ref{lem:lbOPT} and \(\frac{\mu}{2(\mu + 4)} < 1\), we obtain \(
C_\opt \geq \frac{\mu}{2(\mu + 4)} \big(\sum_{t = 1} ^ {T} f_{t}(\bx_t^\star) + \P_{2, T} ^\star + \norm{\bx^\star_{1}} ^ {2}\big).
\)
From Lemma \ref{lem:upOMGD}, we have \begin{align}\label{take_lim_to_0}	C_{\cA_o} \leq \sum_{t = 1} ^ {T} f_{t}(\bx_t^\star) + \frac{1}{2\alpha} \sum_{t = 1} ^ {T} \norm{\nabla f_{t}(\bx_t^\star)} ^ {2} + (L + \alpha + 5)(\norm{\x_{1} - \bx^\star_{1}} ^ 2 + 2\P_{2, T} ^\star  ).\end{align} Since \(\bx_t^\star \in \text{int}(\mathcal{X})\), we have \(\nabla f_{t}(\bx_t^\star) = \mathbf{0}\). Taking \(\lim \alpha \to 0\) in \eqref{take_lim_to_0}, we obtain \begin{align}
	C_{\cA_o} \leq \sum_{t = 1} ^ {T} f_{t}(\bx_t^\star) + (L + 5)(\norm{\bx^\star_{1}} ^ {2} + 2 \P_{2, T} ^ \star) \leq (2L + 10)\big(\sum_{t = 1} ^ {T} f_{t}(\bx_t^\star) + \norm{\bx^\star_{1}} ^ {2} + \P_{2, T} ^\star \big). \nn
\end{align}
Taking the ratio of the obtained bounds on \(C_{\cA_o}\) and \(C_\opt\) completes the proof.
\end{proof}

\begin{rem}\label{rem:comparator}
In this remark, we generalize the competitive ratio attained by the OMGD algorithm \(\cA_o\) against the \(\opt\) \eqref{prob:intro}, to an arbitrary comparator sequence chosen by an user. Formally, let \(\u = (\u_{1}, \dots, \u_{T})\) be an user-specified comparator sequence. The competitive ratio of an online algorithm \(\cA\) that takes an action \(\x_{t}\) at time \(t\), with respect to \(\u\) is defined as \begin{align} \label{generalized_comp_ratio}
	\text{cr}_{\cA} ^ \u := \frac{\sum_{t = 1} ^ T f_{t}(\x_{t}) + c(\x_{t}, \x_{t - 1})}{\sum_{t = 1} ^ T f_{t}(\u_{t}) + c(\u_{t}, \u_{t - 1})}.
\end{align}
Note that if the comparator sequence corresponds to the sequence of optimal actions \((\x_{1}^\opt, \dots, \x_{T}^\opt)\) chosen by the \(\opt\), \eqref{generalized_comp_ratio} corresponds to the competitive ratio against the \(\opt\) \eqref{prob:intro}. When the switching cost is quadratic, from the proof of Theorem \ref{thm:cromgd}, we have \(C_{\cA_o} \leq (2L + 10)\big(\sum_{t = 1} ^ T f_{t}(\bx_t^\star) + \norm{\bx^\star_{1}} ^ {2} + \P_{2, T} ^\star\big)\). Therefore, \(\text{cr}_{\cA_o} ^ {\u}\) can be upper bounded by \begin{align}\nn
	\text{cr}_{\cA_o} ^ {\u} \leq (2L + 10) \frac{ \sum_{t = 1} ^ T f_{t}(\bx_t^\star) + \norm{\x_{t}^\star - \x_{t - 1} ^ \star} ^ 2}{\sum_{t = 1} ^ T f_{t}(\u_{t}) + \frac{1}{2}\norm{\u_{t} - \u_{t - 1}} ^ 2} \stackrel{(a)}\leq 4(L + 5) \bigg(1 + \frac{4}{\mu}\bigg),
\end{align}
where we have defined \(\x_0 ^ \star \coloneqq \mathbf{0}\) and \((a)\) follows from \cite[Thm. 2]{zhang2021revisiting}, which states that \begin{align*}
	\frac{ \sum_{t = 1} ^ T f_{t}(\bx_t^\star) + \frac{1}{2}\norm{\x_{t}^\star - \x_{t - 1} ^ \star} ^ 2}{\sum_{t = 1} ^ T f_{t}(\u_{t}) + \frac{1}{2}\norm{\u_{t} - \u_{t - 1}} ^ 2} \le 1 + \frac{4}{\mu},
\end{align*} for any \(\u\). Therefore, our result on the total cost \(C_{\cA_o}\) attained by OMGD, naturally allows us to generalize the notion of competitive ratio to arbitrary comparator sequences.  
\end{rem}
\textbf{Discussion}: It is instructive to compare our result in Theorem \ref{thm:cromgd} with that of algorithms proposed in  \cite{goel2019online}, called OBD, and \cite{goel2019beyond}, called Regularized OBD, under the full information setting. When all functions $f_t$'s are assumed to be strongly convex with strong convexity parameter $\mu$, the competitive ratio of OBD is at most \(3 + \mathcal{O}(\frac{1}{\mu})\)  \cite{goel2019online} while the competitive ratio of Regularized OBD is at most $\frac{1}{2}\big(1 + \sqrt{1 + \frac{4}{\mu}}\big)$ \cite{goel2019beyond}.
Note that both these algorithms require full information about the function \(f_{t}\) before choosing action $\bx_t$ for their implementation. 
A lower bound of $\Omega(\frac{1}{\sqrt{\mu}})$ as \(\mu \to 0\) on the competitive ratio of all algorithms under the full information setting has also been derived in \cite{goel2019beyond}, showing that Regularized OBD is orderwise optimal. 

Our result in Theorem \ref{thm:cromgd} shows that as \(\mu \to 0\), when $L= c \mu$, {for some finite constant \(c \geq 1\)} (that is independent of \(T\) and all other problem parameters), the competitive ratio of OMGD is bounded by \(\mathcal{O}(\frac{1}{\mu})\) which matches the bound of OBD \cite{goel2019online} which is surprising, since OBD requires full information while OMGD works with limited information. As a function of $\mu$, there is an orderwise gap between the lower bound $\Omega(\frac{1}{\sqrt{\mu}})$ \cite{goel2019beyond} that is applicable under the full information setting and the competitive ratio of OMGD. Therefore, a natural question is: can the lower bound be improved in the limited information setting. 
In the next section, we show that the upper bound obtained in Theorem \ref{thm:cromgd} is orderwise tight for the OMGD algorithm.

Also note that Theorem \ref{thm:cromgd} shows that the competitive ratio of OMGD algorithm depends on both $L, \mu$ in contrast to OBD or Regularized OBD whose bounds only depend on $\mu$. This is a manifestation of the limited information setting. In the next section, we also show that the competitive ratio of any algorithm scales linearly with $L$ in the limited information setting.

\subsection{Lower Bounds on the Competitive Ratio}

We begin by showing that the upper bound on the competitive ratio of the OMGD algorithm obtained in Theorem \ref{thm:cromgd} is order-wise (with respect to \(L, \mu\)) tight.
\begin{lemma}\label{lb:omgd}
\textit{There exists a sequence of functions that satisfy assumption \ref{function_structure}, such that the competitive ratio of the OMGD algorithm is lower bounded by} \begin{align*}
	\text{cr}_{\cA_o} = \frac{C_{\mathcal{A}_{o}}}{{C_\opt}} \geq 1 + \frac{L + 1}{4\mu} + \frac{L + 1}{8}.
\end{align*}
\end{lemma}
The proof of Lemma \ref{lb:omgd} in   \ref{app:lb:omgd} in fact shows that any algorithm, that at time $t$, {\it chases} the optimizer $\bx_{t-1}^\star$ has the same order-wise competitive ratio lower bound as given by Lemma \ref{lb:omgd}. By chasing  the optimizer $\bx_{t-1}^\star$ at time $t$, we mean that 
$||\bx_{t} - \bx_{t-1}^\star|| < c||\bx_{t - 1}- \bx_{t-1}^\star||$ for some constant $0<c<1$ that is independent of all parameters of the problem instance.  Hence, we actually get the following result.
\begin{lemma}\label{lb:chasingalg}
\textit{In the limited information setting, there exists a sequence of functions that satisfy assumption \ref{function_structure}, such that the competitive ratio of any algorithm \(\cA\) that at time $t$, {\it chases} the optimizer $\bx_{t-1}^\star$ is lower bounded by} \begin{align*}
	\text{cr}_{\cA} = \frac{C_{\mathcal{A}}}{{C_\opt}} \geq  \Omega\bigg(\frac{L }{\mu} + L + \frac{1}{\mu}\bigg).
\end{align*}
\end{lemma}
Considering the problem instance used to derive Lemma \ref{lb:omgd},  it is clear that to avoid the competitive ratio lower bound of Lemma \ref{lb:omgd} $\Omega(\frac{L}{\mu})$ when $\mu$ is small or $L$ is large, an algorithm has to move slowly (making it a function of $\mu$ or $L$)
at time $t=2$ from its current position $\bx_1=\mathbf{0}$. This observation gives rise to the consideration of the following class of {\it slow} algorithms that satisfy the property \(||\x_{t} - \x_{t - 1}^\star|| > ||\x_{t - 1} - \x_{t - 1}^\star|| - \epsilon_t\), where $\epsilon_t = \cO(\mu)$ or $\mathcal{O}\left(\frac{1}{L}\right)$ for all $t$, where $\mu$ is small and $L$ is large.
Since we are interested in bounds depending on $\mu$ and $L$, and $L\ge \mu$ by definition, the regime of interest for lower bounds are $\mu\rightarrow 0$ and $L\rightarrow \infty$.  We next show that slow algorithms cannot do better than OMGD in terms of the competitive ratio.

\begin{lemma}\label{lb:slowalg}
\textit{In the limited information setting, there exists a sequence of functions that satisfy assumption \ref{function_structure}, such that the competitive ratio of any slow algorithm  \(\cA\)  is lower bounded by} \[		\text{cr}_{\cA} = \frac{C_{\mathcal{A}}}{{C_\opt}} \geq  \max\big\{\Omega\left(\frac{1}{\mu}\right), \Omega\left(L^2\right)\big\}.
\]

\end{lemma}
The proof of Lemma \ref{lb:slowalg} can be found in   \ref{app:lbslow}. Note that the union of $\x_{t-1}^\star$ chasing or slow algorithms does not cover the class of all algorithms. Next, we present universal lower bounds that apply for all algorithms. We begin by showing that with the information setting as in assumption \ref{ass:info}, the competitive ratio of any  online algorithm  is \(\Omega(L)\).
\begin{lemma}\label{preliminary_lower_bound}
\textit{Under assumptions \ref{ass:info}, \ref{function_structure}, the competitive ratio of any  online algorithm is at least $L$.}
\end{lemma}
The proof of Lemma \ref{preliminary_lower_bound} can be found in  \ref{app:lb1}. The \(\Omega(L)\) bound in Lemma \ref{preliminary_lower_bound} can be improved to $\Omega\big(\frac{L}{\sqrt{\mu}}\big)$ as \(\mu \to 0\), which is stated by the following Lemma.

\begin{lemma}\label{modified_preliminary_lower_bound}
\textit{Under assumptions \ref{ass:info}, \ref{function_structure}, the competitive ratio of any  online algorithm is at least $\Omega\big(\frac{L}{\sqrt{\mu}}\big)$ as \(\mu \to 0\).}
\end{lemma}
The proof of Lemma \ref{modified_preliminary_lower_bound} can be found in  \ref{app:lb2} and follows by constructing an input sequence for which the cost of any online algorithm \(\cA\) is \(C_{\cA} = \Omega(L)\), while \(C_{\opt} = \mathcal{O}(\sqrt{\mu})\) as \(\mu \to 0\). Finally, combining Lemma \ref{preliminary_lower_bound} and Lemma \ref{modified_preliminary_lower_bound}, we get the \(\max\{\Omega(L), \Omega(\frac{L}{\sqrt{\mu}})\}\) lower bound on the competitive ratio of any online algorithm \(\cA\).

\subsection{Dynamic Regret}\label{sec:regret}
In this section, we show that the OMGD algorithm attains a near optimal dynamic regret \eqref{probR:intro}  (order-wise with respect to the path length) under the limited information setting. In order to upper bound the dynamic regret, we assume, similar to \cite{li2020online} that \(f_{t}\)'s have bounded gradients {however, relax the assumption $\x_t ^ \star \in \text{int}(\mathcal{X})$ that was required for deriving Theorem \ref{thm:cromgd}.}
\begin{assumption}\label{bounded_gradients}
For all \(1 \leq t \leq T\), there exists a finite constant \(G < \infty\), such that \(\max_{\x \in \mathcal{X}}\norm{\nabla f_{t}(\x)} = G\).
\end{assumption}
With a generic lower bound on \(C_{\opt}\) \eqref{actual_lower_bound} under our belt and the contraction \(\norm{\x_{t} - \x_{t - 1}^{\star}} \leq \frac{1}{2}\norm{\x_{t - 1} - \x_{t - 1}^\star}\) \eqref{one_fourth_0} attained by OMGD with \(K = \ceil{\frac{L + \mu}{2\mu}\ln 4}\), it is easy to bound the dynamic regret attained by the OMGD algorithm. As usual, we shall assume \(\x_{0} = \mathbf{0}\) as the starting point of all algorithms.
\begin{theorem}\label{regret_omgd}
\textit{Under assumptions \ref{function_structure}, \ref{bounded_gradients}, the dynamic regret  of \(\mathcal{A}_{o}\) (Algorithm \ref{alg:omgd}) with \(K = \ceil{\frac{L + \mu}{2\mu}\ln 4}\) is upper bounded by \[
	R_d(\mathcal{A}_o) \leq 2G( \norm{\x_{1}^\star} + \mathcal{P}_{T}^{\star}) + 5 \norm{\x_{1}^{\star}}^{2} + \bigg(10 - \frac{\mu}{2(\mu + 4)}\bigg) \mathcal{P}_{2, T}^\star.
	\]
} 
\end{theorem}
The proof of Theorem \ref{regret_omgd} is similar to the proof of Theorem \ref{thm:cromgd} and can be found in   \ref{app:regret_thm}. 
The following Corollary shows that when the non-empty closed convex set \(\mathcal{X}\) has a finite diameter \(D\), where \(\max_{\x, \y \in \mathcal{X}} \norm{\x - \y} = D\), the dynamic regret \({R}_{d}(\mathcal{A}_o)\) of the OMGD algorithm is \(\mathcal{O}\big(\norm{\x_{1}^\star} + \mathcal{P}_{T}^\star\big)\).
\begin{corollary}\label{cor:regert_omgd}
\textit{Under assumptions \ref{function_structure}, \ref{bounded_gradients}, and \(\mathcal{X}\) is a non-empty closed convex set with a finite diameter \(D\), the dynamic regret of \(\mathcal{A}_{o}\) is upper bounded by \[
	R_d(\mathcal{A}_o) \leq \bigg(2G + D\bigg( 10 - \frac{\mu}{2(\mu + 4)}\bigg)\bigg)(\norm{\x_{1}^\star} + \mathcal{P}_T^\star).
	\]
}
\end{corollary} 
The proof of Corollary \ref{cor:regert_omgd} follows from Theorem \ref{regret_omgd} by relating \(\mathcal{P}_{2, T}^\star\) with \(\mathcal{P}_{T}^\star\) and can be found in   \ref{app:cor_regret}. In \cite{li2020online}, the {\it delayed information} setting is considered, where at time \(t\), an online algorithm \(\mathcal{A}\) takes action \(\x_{t}\) with complete information about  \(f_{1}, \dots, f_{t - 1}\).
In comparison to our analysis, \cite{li2020online} showed that the online gradient descent algorithm that takes only one gradient step under the delayed information setting achieves a regret of $\cO\left(\frac{G}{1 - \sqrt{1 - \mu/L}} \big(1+\frac{1}{L}\big)(\cP_T^\star + D)\right)$. Thus, when $L \to 0$ and \(\frac{L}{\mu}\) is finite, our bound is superior than \cite{li2020online}. This will also reflect in Theorem \ref{thm:regretomgd} where we show that the OMGD algorithm achieves the order-wise (with respect to the path length) optimal regret.

{A lower bound on the dynamic regret   has been derived in \cite{li2020online} for the {\it delayed information} setting. Given that the limited information setting is weaker than the delayed information setting, this bound is valid for the limited information setting as well, which we state next.}
\begin{lemma}\cite[Thm. 1]{li2020online}\label{fundamental_thm}
\textit{Consider \(T \geq 1\), a closed convex set \(\mathcal{X}\) with a finite diameter \(D\), and a constant \(V_{T}\) s.t. \(0 \leq V_{T}\leq DT\).  For any deterministic online algorithm \(\mathcal{A}\) for \eqref{eq:cost_quadratic}  under the delayed information setting, there exists a sequence of quadratic functions \(\{f_{t}\}_{t = 1}^{T}\) that are \(\mu\)-smooth and \(\mu\)-strongly convex on \(\mathbb{R}^{d}\), satisfy \(\max_{\x \in \mathcal{X}}\norm{\nabla f_{t}(\x)} \leq (3\mu + 1), \) for all \(1 \leq t \leq T\), and \(\norm{
		\x_{1}^\star} + \mathcal{P}_{T}^\star \leq V_{T}\), s.t.
	\begin{align}
		R_d(\mathcal{A}) \geq \zeta DV_{T} \geq \zeta D(\norm{\x_{1}^\star} + \mathcal{P}_{T}^\star),
	\end{align}
}
where \(\zeta = \frac{\mu ^ {3}(1 - \rho)^{2}}{32(\mu + 1)^{2}}, \rho = \frac{\sqrt{\mu + 4} - \sqrt{\mu}}{\sqrt{\mu + 4} + \sqrt{\mu}}\).
\end{lemma}
Thus, we get the main result of this section.
\begin{theorem}\label{thm:regretomgd} \textit{The dynamic regret of the OMGD algorithm is order-wise (with respect to the path length) optimal in the limited information setting when assumptions \ref{function_structure}, \ref{bounded_gradients} hold, and  \(\mathcal{X}\) is a non-empty closed convex set with a finite diameter \(D\).}
\end{theorem}
\begin{proof}
The proof follows by combining Corollary \ref{cor:regert_omgd} and Lemma \ref{fundamental_thm}.
\end{proof}

{It is also interesting to study the regret bound of the OMGD algorithm under the regime when \(\mathcal{X}\) is a non-empty closed convex set and \(\x_{t}^\star \in \text{int}(\mathcal{X})\), but $\cX$ is not necessarily bounded.
In this case,  \(R_d(\mathcal{A}_o)\) can be bounded as \begin{align}
	R_d(\mathcal{A}_o) \leq \big(2L + 10 - \frac{\mu}{2\mu + 8}\big) V_{T} ^ {2},\label{vt_sq_regret}
\end{align}
as \(V_{T} \to 0^{+}\). The proof of \eqref{vt_sq_regret} and discussion regarding \(V_{T} \to 0^{+}\) is provided in  \ref{app:vt_sq_regret}. Comparing \eqref{vt_sq_regret} with Lemma \ref{fundamental_thm}, we can conclude that the lower bound on the dynamic regret derived in Lemma \ref{fundamental_thm} ($\Omega(V_T)$) can be avoided in case $\cX$ does not have a finite diameter and \(\x_{t}^\star \in \text{int}(\mathcal{X})\) which is always true if $\cX=\bbR^d$. To derive Lemma \ref{fundamental_thm}, the construction of `bad' functions \(\{f_{t}\}_{t = 1}^{T}\) was such that the minimizer \(\x_{t}^\star\) lied on the boundary of \(\mathcal{X}\) and had a non-zero gradient \(\norm{\nabla f_{t}(\x_{t}^\star)} \neq 0\). Thus, a different lower bound is necessitated when $\cX$ does not have a finite diameter and \(\x_{t}^\star \in \text{int}(\mathcal{X})\), compared to Lemma \ref{fundamental_thm}.}

\section{Linear Switching Cost}
In this section, we consider \eqref{cost:intro} in the context of the linear switching cost, i.e., the cost  \eqref{cost:intro} of an online algorithm \(\cA\) is given by \begin{align}\label{eq:cost_linear}
C_{\cA} = \sum_{t = 1} ^ T f_{t}(\x_{t}) + \norm{\x_{t} - \x_{t - 1}}.
\end{align}Most of the results in this section are motivated from the derived results for quadratic switching cost.
\subsection{Competitive Ratio}

\subsubsection{\textup{Upper Bounding the Cost \eqref{eq:cost_linear} of OMGD\nopunct}\\} 
Compared to \eqref{eq:cost_quadratic}, \eqref{eq:cost_linear} differs only in the switching cost. Thus, to analyze OMGD  \((\cA_o)\) with linear switching cost, we have to 
separately bound the total switching cost, which we do as follows  similar to Lemma \ref{lemma_switching_cost}.
\begin{lemma}\label{lemma_switching_cost_linear}
\textit{Under assumption \ref{function_structure}, for \(\cA_o\) with \(K = \ceil{\frac{L + \mu}{2\mu}{\ln 4}}\), the total switching cost is bounded by \begin{align*}
		\sum_{t = 2} ^ T \norm{\x_{t} - \x_{t - 1}} \leq 3 \norm{\x_{1} - \x_{1}^\star} + 3 \cP_{T}^\star.
\end{align*}}
\end{lemma}
\begin{proof}
We begin by bounding the quantity \(\sum_{t = 1}^{T} \norm{\x_{t} - \x_{t}^\star}\) for \(\mathcal{A}_o\) in the following manner:
\begin{align}
	\sum_{t = 2} ^ {T} \norm{\x_{t} - \x_{t}^\star} & \stackrel{(a)}\leq \sum_{t = 2} ^ {T} \big(\norm{\x_{t} - \x_{t - 1}^\star} + \norm{\x_{t}^\star - \x_{t - 1}^\star}\big), \nn\\
	& \stackrel{(b)}\leq \sum_{t = 2} ^ {T} \frac{1}{2}\norm{\x_{t - 1} - \x_{t - 1}^\star} + \mathcal{P}_{T}^{\star}, \nn\\
	&= \frac{1}{2}\norm{\x_{1} - \x_{1}^\star} + \sum_{t = 2} ^ {T - 1} \frac{1}{2}\norm{\x_{t} - \x_{t}^\star} + \mathcal{P}_{T}^\star, \nn \\
	& \leq \frac{1}{2}\norm{\x_{1} - \x_{1}^\star} + \sum_{t = 2} ^ {T} \frac{1}{2}\norm{\x_{t} - \x_{t}^\star} + \mathcal{P}_{T}^\star, \nn
\end{align}
where \((a)\) follows from the Triangle inequality, while \((b)\) follows from the contraction bound \eqref{one_fourth_0} attained by \(\mathcal{A}_{o}\). Rearranging, we obtain \begin{align}
	\sum_{t = 2} ^ {T} \norm{\x_{t} - \x_{t}^\star} \leq \norm{\x_{1} - \x_{1}^\star} + 2\mathcal{P}_{T}^\star \implies \sum_{t = 1} ^ {T} \norm{\x_{t} - \x_{t}^\star} \leq 2\norm{\x_{1} - \x_{1}^\star} + 2\mathcal{P}_{T}^\star. \label{cumulative_sum_cr_linear}
\end{align}
The overall switching cost of \(\cA_o\) can therefore be bounded as \begin{align*}
	\sum_{t = 1}^{T} \norm{\x_{t} - \x_{t - 1}} &\stackrel{(a)}= \sum_{t = 2}^{T} \norm{\x_{t} - \x_{t - 1}}\stackrel{(b)}\leq \sum_{t = 2}^{T} \big( \norm{\x_{t} - \x_{t - 1}  ^ \star} + \norm{\x_{t - 1} - \x_{t - 1} ^ \star} \big)\stackrel{(c)}\leq \frac{3}{2} \sum_{t = 1}^{T - 1}\norm{\x_{t} - \x_{t}^\star}, \\
	&\stackrel{(d)}\leq 3\norm{\x_{1} - \x_{1}^\star} + 3\cP_{T}^\star,
\end{align*}
where \((a)\) follows since \(\x_{1} = \x_0\), \((b)\) follows from the Triangle inequality, \((c)\) follows from \eqref{one_fourth_0}, and \((d)\) follows from \eqref{cumulative_sum_cr_linear}, derived above.
\end{proof}
Finally, combining Lemma \ref{lemma_hitting_cost} (total hitting cost of \(\cA_o\)) and Lemma \ref{lemma_switching_cost_linear} (total switching cost of \({\cA_o}\)), we get a bound on the total cost \eqref{eq:cost_linear} attained by \(\cA_o\).
\begin{lemma}\label{up:omgd_linear}
\textit{Under assumption \ref{function_structure}, the total cost of $\cA_o$ \eqref{eq:cost_linear} with \(K = \ceil{\frac{L + \mu}{2\mu}\ln 4}\) is bounded by \[
	C_{\cA_o} \leq \sum_{t = 1} ^ {T} f_{t}(\x_t^\star) + \frac{1}{2\alpha} \sum_{t = 1} ^ {T} \norm{\nabla f_{t}(\x_t^\star)} ^ {2} + (L + \alpha)(\norm{\x_{1} - \x^\star_{1}} ^ 2 + 2\P_{2, T} ^\star  ) + 3\norm{\x_{1} - \x_{1} ^ \star} + 3\cP_{T}^\star,\]
	for any \(\alpha > 0\).}	
\end{lemma}

Observe from Lemma \ref{up:omgd_linear} that the total cost \(C_{\cA_o}\) depends on both \(\cP_{2, T} ^ \star\) and \(\cP_{T} ^ \star\), unlike the quadratic switching cost case where the total cost, given by Lemma \ref{lem:upOMGD}, depends only on \(\cP_{2, T}^\star\). This is a manifestation of the linear switching cost. In the next section, we show that the total cost of the \(\opt\), \(C_\opt\), also depends on both \(\cP_{2, T}^\star\) and \(\cP_{T}^\star\).

\subsubsection{{\textup{Lower Bounding the cost of the \(\opt\)\nopunct}}\\}
In this section, we derive a lower bound on \(C_{\opt}\) for the linear switching cost problem by reducing the linear switching cost problem to a particular instance of the quadratic switching cost problem, and then reusing
the  lower bound on \(C_{\opt}\) for the quadratic switching cost problem (Lemma \ref{lem:lbOPT}). Similar to the quadratic switching cost case, we assume \(\x_0 = \mathbf{0}\) as the starting point of all algorithms.

\ \begin{lemma}\label{lem:lbOPT_linear}
\textit{Under assumption \ref{function_structure}, the cost incurred by the $\opt$ \eqref{eq:cost_linear} satisfies \begin{align}\label{actual_lower_bound_linear}
		C_{\opt} &\geq \sum_{t = 1} ^ {T} f_{t}(\x_{t}^\star) + \frac{2\mu (\cP_{2, T} ^ \star + \norm{\x_{1}^\star} ^ {2})}{\mu (\cP_{T} ^ \star + \norm{\x_{1} ^\star}) + 8}.
	\end{align}
}
\end{lemma}
\begin{proof}
We begin by lower bounding \(C_{\opt}\) in the following manner: 
\begin{align}
	C_{\opt} &= \min_{\{\x_{t} \in \cX\}_{t = 1} ^ {T}} \sum_{t = 1} ^ {T}\big(f_{t}(\x_{t}) + \norm{\x_{t} - \x_{t - 1}}\big),\nn \\
	&\stackrel{\ref{function_structure}}\geq  \min_{\{\x_{t} \in \mathcal{X}\}_{t = 1} ^ {T}} \sum_{t = 1} ^ {T} \big(f_{t}(\bx_t^\star) + \ip{\nabla f_{t}(\bx_t^\star)}{\x_{t} - \bx_t^\star} + \frac{\mu}{2} \norm{\x_{t} - \bx_t^\star} ^ {2} + \norm{\x_{t} - \x_{t - 1}}\big)\nn, \\
	&\stackrel{(a)}\geq \sum_{t = 1} ^ {T} f_{t}(\x_{t} ^ {\star}) + \min_{\{\x_{t} \in \cX\}_{t = 1} ^ {T}}\sum_{t = 1} ^ {T} \big(\frac{\mu}{2}\norm{\x_{t} - \x_{t}^\star} ^ {2} + \norm{\x_{t} - \x_{t - 1}}\big),\nn \\
	&\stackrel{(b)}= \sum_{t = 1} ^ {T} f_{t}(\x_{t} ^ \star) + \sum_{t = 1} ^ {T} \big( \frac{\mu}{2}\norm{\x_{t} ^ {\omega} - \x_{t} ^ \star} ^ {2} + \norm{\x_{t}^\omega - \x_{t - 1}^\omega}\big),\nn \\
	&\stackrel{(c)}\geq \sum_{t = 1} ^ {T} f_{t}(\x_{t}^\star) + \sum_{t = 1} ^ {T} \big(\frac{\mu}{2}\norm{\x_{t} ^ \omega - \x_{t}^\star} ^ {2} + \frac{1}{2U}\norm{\x_{t} ^ {\omega}- \x_{t - 1} ^ \omega}^{2}\big),\nn \\
	&\geq \sum_{t = 1} ^ {T} f_{t}(\x_{t}^\star) + \min_{\{\x_{t} \in \cX\}_{t = 1} ^ {T}}\sum_{t = 1} ^ {T} \big(\frac{\mu}{2}\norm{\x_{t} - \x_{t}^\star} ^ {2} + \frac{1}{2U}\norm{\x_{t} - \x_{t - 1}}^{2}\big),\nn \\
	&\stackrel{(d)}\geq \sum_{t = 1} ^ {T} f_{t}(\x_{t}^\star) + \frac{\mu}{\mu U + 4}\big(\cP_{2, T} ^ \star + \norm{\x_{1}^\star}^{2}\big),\label{continue_from_here}
\end{align}
where \((a)\) follows from the optimality condition of \(\x_{t}^\star = \argmin_{\x \in \cX} f_{t}(\x)\), i.e., \(\ip{\nabla f_{t}(\x_{t}^\star)}{\x - \x_{t}^\star} \ge 0\), for all \(\x \in \cX\). To get \((b)\), we have defined \[\{\x_{t} ^ {\omega}\}_{t = 1} ^ {T} := \min_{\{\x_{t} \in \cX\}_{t = 1} ^ {T}} \sum_{t = 1} ^ {T} \big(\frac{\mu}{2}\norm{\x_{t} - \x_{t}^\star} ^ {2} + \norm{\x_{t} - \x_{t - 1}}\big),\] and in \((c)\) 
$U$ is such that \(\norm{\x_{t}^\omega - \x_{t - 1}^\omega} \le 2U\), for all \(1 \leq t \leq T\). To obtain \((d)\), we have used the result of Lemma \ref{lem:lbOPT}.
It only remains to evaluate \(U\) which we do as follows.

Since \(\{\x_{t} ^ {\omega}\}_{t = 1} ^ {T} = \min_{\{\x_{t} \in \cX\}_{t = 1} ^ {T}} \sum_{t = 1} ^ {T} \big(\frac{\mu}{2}\norm{\x_{t} - \x_{t}^\star} ^ {2} + \norm{\x_{t} - \x_{t - 1}}\big)\), we have the following: \begin{align}
	\sum_{t = 1} ^ {T} \big(\frac{\mu}{2}\norm{\x_{t}^{\omega} - \x_{t}^\star} ^ {2} + \norm{\x_{t}^{\omega} - \x_{t - 1}^{\omega}}\big) \leq \cP_{T}^\star + \norm{\x_{1}^\star}\label{useful_step},
\end{align}
which follows by setting \(\x_{t} = \x_{t}^\star\) in the minimization \( \min_{\{\x_{t} \in \cX\}_{t = 1} ^ {T}} \sum_{t = 1} ^ {T} \big(\frac{\mu}{2}\norm{\x_{t} - \x_{t}^\star} ^ {2} + \norm{\x_{t} - \x_{t - 1}}\big) \). From \eqref{useful_step}, we obtain the following: \begin{align}
	\sum_{t = 1} ^ {T} \norm{\x_{t} ^ \omega - \x_{t - 1} ^ \omega} \leq \cP_{T} ^ \star + \norm{\x_{1}^\star} \implies \norm{\x_{t} ^ \omega - \x_{t - 1} ^ \omega} \leq \cP_{T}^\star + \norm{\x_{1}^\star}\nn,
\end{align}
for all \(1 \leq t \leq T\). Therefore, it is sufficient to choose \(2U = \cP_{T}^\star + \norm{\x_{1}^\star}\). Using this value of $U$ in  \eqref{continue_from_here}, we get \begin{align}
	C_{\opt} &\geq \sum_{t = 1} ^ {T} f_{t}(\x_{t}^\star) + \frac{2\mu (\cP_{2, T} ^ \star + \norm{\x_{1}^\star} ^ {2})}{\mu (\cP_{T} ^ \star + \norm{\x_{1} ^\star}) + 8}. \nn
\end{align}
which completes the proof.
\end{proof}
\subsubsection{{\textup{Bounding the Competitive Ratio of OMGD\nopunct}}\\}
Combining Lemma \ref{up:omgd_linear} and Lemma \ref{lem:lbOPT_linear}, we can bound the competitive ratio of OMGD(\(\cA_o\)) for the linear switching cost case, for any value of \(f_{t}(\x_{t}^\star)\). However, to get a simplified expression we consider \(f_{t}(\x_t ^ \star) = 0\) and derive the following result.
\begin{theorem}\label{thm:cromgd_linear}
\textit{Under assumption \ref{function_structure}, and assuming that the minimizers \(\bx_t^\star \in \text{int}(\mathcal{X})\) and \(f_{t}(\x_{t}^\star) = 0\), the competitive ratio of the OMGD algorithm $\cA_o$ for the linear switching cost satisfies} \begin{align}\label{actual_upper_bound_cr}
	\text{cr}_{\cA_o} = \frac{C_{\cA_o}}{C_{\opt}}\leq L(\cP_{T}^\star + \norm{\x_{1}^\star}) + \frac{3}{2} \frac{\big(\cP_{T}^\star + \norm{\x_{1}^\star}\big) ^ 2}{\cP_{2, T}^\star + \norm{\x_{1}^\star} ^ 2} + \frac{8L}{\mu} + \frac{12}{\mu} \frac{\cP_{T}^\star + \norm{\x_{1}^\star}}{\cP_{2, T}^\star + \norm{\x_{1}^\star} ^ 2}.
\end{align}
\end{theorem}
\begin{proof}
From Lemma \ref{lem:lbOPT_linear}, we have \(C_{\opt} \ge \frac{2\mu(\cP_{2, T}^\star + \norm{\x_{1}^\star} ^ 2)}{\mu(\cP_{T}^\star + \norm{\x_{1}^\star}) + 8}\). Further, since \(\x_{t}^\star \in \text{int}(\cX)\) and \(f_{t}(\x_{t}^\star) = 0\), letting \(\alpha \to 0\) in Lemma \ref{up:omgd_linear}, we have \(C_{\cA_o} \le 2L(\norm{ \x_{1}^\star} ^ 2 + \cP_{2, T}^\star) + 3(\norm{\x_{1}^\star} + \cP_{T}^\star)\). Taking the ratio of both the bounds completes the proof.
\end{proof}

Observe from Theorem \ref{thm:cromgd_linear}, the competitive ratio of \(\cA_o\) depends on \(\cP_{2, T}^\star\) and \(\cP_{T}^\star\), in addition to \(L, \mu\). This is in contrast to the competitive ratio of \(\cA_o\) for the quadratic switching cost (Theorem \ref{thm:cromgd}), which solely depends on \(L, \mu\). Therefore, it is natural to consider whether the dependence of \(\cP_{2, T}^\star\) and \(\cP_{T}^\star\) in the upper bound of OMGD can be avoided. In the next section, we show that this dependence cannot be avoided, by constructing a problem instance where the lower bound for any online algorithm \(\cA\) depends on both \(\cP_{2, T}^\star\) and \(\cP_{T}^\star\).
\subsection{Lower Bound on the Competitive Ratio}

To explicitly bring out the dependence of the lower bound for any online algorithm \(\cA\) on \(\cP_{2, T}^\star\) and \(\cP_{T}^\star\), we assume \(\mu = L\) for simplicity. The following Lemma shows that there exists a problem instance where the explicit dependence of \(\cP_{2, T}^\star\) and \(\cP_{T} ^ \star\) in the lower bound matches with the upper bound of OMGD (Theorem \ref{thm:cromgd_linear}). The proof of Lemma \ref{lem:lb_linear} can be found in \ref{app:lb_linear} and proceeds by creating two functions \(f_{1}, f_{2}\) s.t. \(f_{1}(x) = \frac{\mu}{2}(x - \theta) ^ 2\), \(f_{2}(x) = \frac{\mu}{2}x^2\). It can be shown that for this choice of \(f_{1}, f_{2}\), any online algorithm \(\cA\) incurs cost \(C_{\cA}\) which is quadratic in \(\theta\) while  \(C_{\opt}\)  is linear in \(\theta\). The ratio of \(C_{\cA}\) and  \(C_{\opt}\) can then be appropriately expressed as $\frac{1}{8}$ times the quantity in \eqref{lb_linear_any_A}.
\begin{lemma}\label{lem:lb_linear}
\textit{In the limited information setting, there exists a sequence of functions that satisfy assumption \ref{function_structure}, such that the competitive ratio of any  online algorithm \(\cA\) for the linear switching cost problem is \begin{align}\label{lb_linear_any_A}\Omega \bigg( \mu(\cP_{T}^\star + \norm{\x_{1}^\star}) + \frac{3}{2} \frac{\big(\cP_{T}^\star + \norm{\x_{1}^\star}\big) ^ 2}{\cP_{2, T}^\star + \norm{\x_{1}^\star} ^ 2} + \frac{12}{\mu} \frac{\cP_{T}^\star + \norm{\x_{1}^\star}}{\cP_{2, T}^\star + \norm{\x_{1}^\star} ^ 2}\bigg).\end{align}}
\end{lemma}

\begin{rem}
In contrast to the quadratic switching cost case, the lower bound (Lemma \ref{lem:lb_linear}) on the competitive ratio of any online algorithm depends on both the path length and the squared path length. An easy explanation for this can be generated by restricting 
the strongly convex functions to the class of strongly convex quadratic functions of the form \(f_{t}(\x) = \frac{\mu}{2}\norm{\x - \x_{t}^\star} ^ 2\). In this case, when the switching cost is linear, choosing
$\bx_t = \bx_t^\star$, $\opt$ only incurs a cost of  
$\cP_{T}^\star$, which however cannot be matched by an online algorithm which ends up incurring a cost that is both a function of $\cP_{T}^\star$ and $\cP_{2, T}^\star$. When the switching cost is quadratic, however, choosing
$\bx_t = \bx_t^\star$, $\opt$ also incurs a cost of  
$\cP_{2, T}^\star$, which is similar to that of an online algorithm.
%
%
%
%
\end{rem}

\section{Conclusion}
In this paper, we considered the OCO-S problem with quadratic and linear switching cost in the limited information setting and proposed an algorithm whose competitive ratio was shown to be order-wise (in terms of \(L, \mu\)) tight when the objective functions are  $L$-smooth and $\mu$-strongly convex, and the switching cost is quadratic. Lower bounds on the competitive ratio of any  online algorithm for the quadratic switching cost are also derived, that show that the linear dependence of $L$ on the competitive ratio is unavoidable. The proposed algorithm is also shown to achieve the optimal order-wise (with respect to the path length) dynamic regret in the limited information setting when the feasible set is bounded and the switching cost is quadratic. We also upper bounded  the dynamic regret of the proposed algorithm when the feasible set is unbounded, and showed that it can be smaller than the lower bound for the case when the feasible set is bounded under a certain regime. Finally, we showed that the linear switching cost problem is fundamentally different from the quadratic switching cost problem, since the dependence of the squared path length and path length on the competitive ratio is unavoidable for any online algorithm in this case.

\bibliographystyle{elsarticle-num}
\bibliography{references, refs}

\newpage
\appendix
\part{Appendices} 
\section{Lemma \ref{lem:contraction}}\label{app:lem:contraction}
The following result has been derived in \cite{zhang2017improved} and is used in Lemma \ref{squared_distance_lemma} for our analysis.
\begin{lemma}\cite[Lemma 5]{zhang2017improved} \label{lem:contraction}
	\textit{Assume that \(f: \mathcal{X} \to \mathbb{R}^{+}\) is \(L\)-smooth and \(\mu\)-strongly convex, and \(\bx^\star = \argmin_{\x \in \mathcal{X}} f(\x)\). Let $\bu \in \cX$ be the current point, and let the new point $\v = \mathcal{P}_{\mathcal{X}}(\u - \eta \nabla f(\u))$ be obtained by using a single gradient descent step with step size $\eta$, where \(\eta \leq \frac{1}{L}\). Then,  \[
		||{\v - \bx^\star}|| ^ {2} \leq \bigg(1 - \frac{2\mu}{\frac{1}{\eta} + \mu}\bigg) ||{\u - \bx^\star}|| ^ {2}. 
		\]
	} 
\end{lemma}
\begin{proof}
	Observe that the update \(\v = \mathcal{P}_{\mathcal{X}}(\u - \eta \nabla f(\u))\) can be expressed as the following: \begin{align}\label{another_way_of_seeing_projection}
		\v = \argmin_{\x \in \mathcal{X}} f(\u) + \ip{\nabla f(\u)}{\x - \u} + \frac{1}{2\eta} \norm{\x - \u} ^ {2}.
	\end{align}
Next, consider a function \(g: \mathcal{X} \to \mathbb{R}^{+}\), which is \(\lambda\)-strongly convex. Denote \(\x_{g}^\star = \argmin_{\x \in \mathcal{X}} g(\x)\), which implies that \begin{align}\label{opt_cond_again}
	\ip{\nabla g(\x_{g}^\star)}{\x - \x_{g}^\star} \geq 0, \forall \x \in \mathcal{X}.
\end{align}
From the \(\lambda\)-strong convexity of \(g\) over \(\mathcal{X}\) we have the following for all \(\x \in \mathcal{X}\): \begin{align}
	g(\x) &\geq g(\x_{g}^\star) + \ip{\nabla g(\x_{g}^\star)}{\x - \x_{g}^\star} + \frac{\lambda}{2} \norm{\x - \x_{g}^\star} ^ {2} \geqtext{\eqref{opt_cond_again}} g(\x_{g}^\star) + \frac{\lambda}{2} \norm{\x - \x_{g}^\star} ^ {2}.\label{hazan_result}
\end{align}
Observe that the minimization in \eqref{another_way_of_seeing_projection} is performed over a function which is $\frac{1}{\eta}$ strongly convex over \(\mathcal{X}\), and has a minimizer \(\v\). Also, recall that \(\x^\star = \argmin_{\x \in \mathcal{X}} f(\x)\). Therefore, from \eqref{hazan_result} with \(\x = \x^\star, g(\x) = f(\u) + \ip{\nabla f(\u)}{\x - \u} + \frac{1}{2\eta} \norm{\x - \u}^{2}, \x^\star_{g} = \v\), we have the following:
\begin{align}
	& f(\u) + \ip{\nabla f(\u)}{\x^\star - \u} + \frac{1}{2\eta} \norm{\u - \x^\star} ^ {2} \geqtext{\eqref{hazan_result}} f(\u) + \ip{\nabla f(\u)}{\v - \u} +  \frac{1}{2\eta} \norm{\v - \u} ^ {2} + \frac{1}{2\eta} \norm{\v - \x^\star} ^ {2} \nn, \\
	\implies & f(\u) + \ip{\nabla f(\u)}{\v - \u} + \frac{1}{2\eta} \norm{\v - \u} ^ {2} \leq f(\u) + \ip{\nabla f(\u)}{\x^\star - \u} + \frac{1}{2\eta} \norm{\u - \x^\star} ^ {2} - \frac{1}{2\eta} \norm{\v - \x^\star} ^ {2}. \label{step}
\end{align}
From the \(\mu\)-strong convexity of \(f\) we have the following: \begin{align}\nn
	f(\u) + \ip{\nabla f(\u)}{\x^\star - \u} \leq f(\x^\star) - \frac{\mu}{2} \norm{\u - \x^\star} ^ {2}.
\end{align}
Using strong convexity to bound \(f(\u) + \ip{\nabla f(\u)}{\x^\star - \u}\) in \eqref{step}, we obtain \begin{align}
	& f(\u) + \ip{\nabla f(\u)}{\v - \u} + \frac{1}{2\eta} \norm{\v - \u} ^ {2} \leq f(\x^\star) + \frac{1}{2\eta} \norm{\u - \x^\star} ^ {2} - \frac{\mu}{2} \norm{\u - \x^\star} ^ {2}  - \frac{1}{2\eta} \norm{\v - \x^\star} ^ {2}, \nn\\
	 \implies &  f(\u) + \ip{\nabla f(\u)}{\v - \u} \leq f(\x^\star) + \frac{1}{2\eta} \norm{\u - \x^\star} ^ {2} - \frac{\mu}{2} \norm{\u - \x^\star} ^ {2} - \frac{1}{2\eta} \norm{\v - \x^\star} ^ {2} - \frac{1}{2\eta} \norm{\v - \u} ^ {2}. \label{intermediate_step_}
\end{align}
The \(L\)-smoothness of \(f\) and \(\eta \leq \frac{1}{L}\) implies \begin{align}
	f(\v) \leq f(\u) + \ip{\nabla f(\u)}{\v - \u} + \frac{L}{2} \norm{\v - \u} ^ {2} \leq f(\u) + \ip{\nabla f(\u)}{\v - \u} + \frac{1}{2\eta}\norm{\v - \u} ^ {2}. \label{another_step}
\end{align}
Using \eqref{intermediate_step_} to bound \(f(\u) + \ip{\nabla f(\u)}{\v - \u}\) in \eqref{another_step}, we obtain the following:
\begin{align}\label{penultimate_step}
	f(\v) \leqtext{\eqref{intermediate_step_}, \eqref{another_step}} f(\x^\star) + \frac{1}{2\eta} \norm{\u - \x^\star} ^ {2} - \frac{\mu}{2} \norm{\u - \x^\star} ^ {2} - \frac{1}{2\eta} \norm{\v - \x^\star} ^ {2}.
\end{align}
Again, applying \eqref{hazan_result} with \(g = f, \x^\star_{g} = \x^\star, \x = \v\), we have \begin{align}\label{hazan_again}
	f(\v) \geqtext{\eqref{hazan_result}} f(\x^\star) + \frac{\mu}{2} \norm{\v - \x^\star} ^ {2}.
\end{align}
Combining \eqref{penultimate_step} and \eqref{hazan_again}, we obtain \begin{align}
	& \frac{\mu}{2} \norm{\v - \x^\star} ^ {2} \leqtext{\eqref{hazan_again}} f(\v) - f(\x^\star) \leqtext{\eqref{penultimate_step}} \frac{1}{2\eta} \norm{\u - \x^\star} ^ {2} - \frac{\mu}{2} \norm{\u - \x^\star} ^ {2} - \frac{1}{2\eta} \norm{\v - \x^\star} ^ {2},\nn \\
 & \hspace{20mm}	\implies\bigg(\frac{1}{\eta} + \mu\bigg) \norm{\v - \x^\star} ^ {2} \leq \bigg(\frac{1}{\eta} - \mu\bigg) \norm{\u - \x^\star} ^ {2}.\nn
\end{align}
This completes the proof.
\end{proof}
\section{Nesterov Accelerated Gradient (NAG)}\label{app:NAG}
In this section, we show that to get \(1/4\)-th contraction as in \eqref{one_fourth_0}, the number of gradient queries can be reduced to \({\cO}(\sqrt{Q}\log Q)\), where \(Q := \frac{L}{\mu}\) is the condition number of \(f\), by the use of NAG. For simplicity, we assume that \(\cX = \Rn ^ d\).

We begin with a brief introduction of NAG. Consider the following optimization problem:\begin{align}
	\x^\star = \argmin_\x f(\x), \tag{\(\cP\)}\label{eq:problem}
\end{align}
where \(f\) is \(\mu\)-strongly convex and \(L\)-smooth. For \(\eqref{eq:problem}\), NAG maintains the sequence \(\{\x_k, \y_k\}\) and is entailed by the following updates:
\begin{align}
	\y_{k + 1} = \x_{k} - \frac{1}{L} \nabla f(\x_{k}), \quad \x_{k + 1} = \bigg(1 - \frac{\sqrt{Q} - 1}{\sqrt{Q} + 1}\bigg) \y_{k + 1} + \frac{\sqrt{Q} - 1}{\sqrt{Q} + 1} \y_{k}.
\end{align}
The following Theorem from \cite{nesterov2003introductory} shows that NAG enjoys a convergence rate of \(\exp\big(-\frac{k}{\sqrt{Q}}\big)\).
\begin{theorem}\cite{nesterov2003introductory} \label{nag_rate}
	\textit{The sequence \(\{\x_{k}, \y_k\}\) produced by NAG, satisfies \begin{align}\label{exp_rate}
		f(\y_{k}) - f(\x^\star) \leq \frac{L + \mu}{2} \norm{\x_0 - \x^\star} ^ 2 \exp (-\frac{k - 1}{\sqrt{Q}}).
	\end{align}
}
\end{theorem}
From Theorem \ref{nag_rate}, we can easily bound the residual \(\norm{\y_{k} - \x^\star}\) by the following:
\begin{align}
	f(\y_{k}) & \stackrel{(a)}\geq f(\x^\star) + \ip{\nabla f(\x^\star)}{ \y - \x^\star} + \frac{\mu}{2} \norm{\y_k - \x^\star} ^ 2 \nn , \\
	&= f(\x ^ \star) + \frac{\mu}{2}\norm{\y_k - \x^\star} ^ 2\label{triv},
\end{align}
where \((a)\) follows from the \(\mu\)-strong convexity of \(f\). Therefore, we have \begin{align}
	&\frac{\mu}{2} \norm{\y_k - \x^\star} ^ 2 \stackrel{\eqref{triv}}\leq f(\y_k) - f(\x^\star) \stackrel{\eqref{exp_rate}}\leq \frac{L + \mu}{2} \norm{\x_0 - \x^\star} ^ 2 \exp (-\frac{k - 1}{\sqrt{Q}}), \nn \\
	& \hspace{10mm}\implies \norm{\y_{k + 1} - \x^\star} ^ 2 \leq  \frac{L + \mu}{\mu} \norm{\x_0 - \x^\star} ^ 2 \exp (-\frac{k}{\sqrt{Q}}) \label{temp_step}. 
\end{align}

We now discuss the resulting counterpart of the OMGD  algorithm. At time \(t\), we start with \(\y_{0} ^ t = \x_{0} ^ t = \x_{t - 1}\) and execute NAG with gradients corresponding to \(\nabla f_{t - 1}\) (and therefore \(\x^\star = \x_{t - 1}^\star\)). The final action is taken as \(\x_{t} = \y_{K + 1} ^ t\), where \(K = \ceil{\sqrt{Q}\ln 4(Q + 1)}\). 
From \eqref{temp_step}, we have the following: \begin{align}
	\norm{\x_{t} - \x_{t - 1}^\star} ^ 2 &\leq \exp{(\ln (Q + 1))}\exp (-\frac{K}{\sqrt{Q}}) \norm{\x_{t - 1} - \x_{t - 1}^\star} ^ 2\nn, \\
	& \leq \exp{\big(\ln(Q + 1) -\frac{K}{\sqrt{Q}}\big)}  \norm{\x_{t - 1} - \x_{t - 1}^\star} ^ 2 \nn, \\
	& \leq \frac{1}{4} \norm{\x_{t - 1} - \x_{t - 1}^\star} ^ 2.\nn
\end{align}
\section{Proof of Lemma \ref{lemma_hitting_cost}}\label{app:hitting_cost}
	From the \(L\)-smoothness of \(f\), as per Assumption \ref{function_structure}, we have \begin{align}
		f_t(\x_{t}) & \leq f_{t}(\bx_t^\star) + \ip{\nabla f_{t}(\bx_t^\star)}{\x_{t} - \bx_t^\star} + \frac{L}{2} \norm{\x_{t} - \bx_t^\star} ^ 2 \nn, \\
		& \stackrel{(a)}\leq f_{t}(\bx_t^\star) + \norm{\nabla f_{t} (\bx_t^\star)} \norm{\x_{t} - \bx_t^\star} + \frac{L}{2} \norm{\x_{t} - \bx_t^\star} ^ 2 \nn, \\
		& \stackrel{(b)}\leq f_{t}(\bx_t^\star) + \frac{1}{2\alpha}\norm{\nabla f_{t}(\bx_t^\star)} ^ 2 + \frac{L + \alpha}{2} \norm{\x_{t} - \bx_t^\star} ^ 2 \nn,
	\end{align}
	where \((a)\) follows by applying the Cauchy-Schwarz inequality \(\ip{\u}{\v} \leq \norm{\u} \norm{\v}\), where \(\u, \v \in \mathbb{R} ^ {d}\), while \((b)\) follows by applying the Peter-Paul inequality \(
	ab \leq \frac{a ^ {2}}{2\alpha} + \frac{\alpha b ^ {2}}{2},
	\)
	which holds for any \(\alpha > 0, a, b \in \mathbb{R}\).
	Thus the total hitting cost of $\cA_o$ can be bounded by \begin{align}
		\sum_{t = 1} ^ {T} f_{t}(\x_{t}) & \leq \sum_{t = 1} ^ {T} f_{t}(\bx_t^\star) + \frac{1}{2\alpha} \sum_{t = 1} ^ {T} \norm{\nabla f_{t}(\bx_t^\star)} ^ 2 + \frac{L + \alpha}{2} \sum_{t = 1} ^ {T} \norm{\x_{t} - \bx_t^\star} ^ 2 \nn, \\
		& \leqtext{\eqref{actual_squared_bound}} \sum_{t = 1} ^ {T} f_{t}(\bx_t^\star) + \frac{1}{2\alpha} \sum_{t = 1} ^ {T} \norm{\nabla f_{t}(\bx_t^\star)} ^ 2 + {(L + \alpha)}(\norm{\x_{1} - \bx^\star_{1}}^{2} + 2 \P_{2, T} ^\star) \nn.
	\end{align}
	This completes the proof.

\section{Proof of Lemma 
	\ref{lem:lbOPT}}\label{app:lbOPT}
\begin{proof}	The cost of the $\opt$ is given by \begin{align}
		C_\opt &= \min_{\{\x_{t} \in \mathcal{X}\}_{t = 1} ^ {T}} \sum_{t = 1} ^ {T} \bigg(f_{t}(\x_{t}) + \frac{1}{2} \norm{\x_{t} - \x_{t - 1}} ^ {2}\bigg),
		\nn \\
		 	  & \stackrel{(a)}\geq  \min_{\{\x_{t} \in \mathcal{X}\}_{t = 1} ^ {T}} \sum_{t = 1} ^ {T} \big(f_{t}(\bx_t^\star) + \ip{\nabla f_{t}(\bx_t^\star)}{\x_{t} - \bx_t^\star} + \frac{\mu}{2} \norm{\x_{t} - \bx_t^\star} ^ {2} + \frac{1}{2} \norm{\x_{t} - \x_{t - 1}} ^ {2} \big)\nn, \\
			  & \stackrel{(b)}\geq \min_{\{\x_{t} \in \mathcal{X}\}_{t = 1} ^ {T}} \sum_{t = 1} ^ {T} \big(f_{t}(\bx_t^\star) + \frac{\mu}{2} \norm{\x_{t} - \bx_t^\star} ^ {2} + \frac{1}{2} \norm{\x_{t} - \x_{t - 1}} ^ {2}\big) \nn,\\
			  & =\sum_{t = 1} ^ {T} f_{t}(\bx_t^\star) + \min_{\{\x_{t} \in \mathcal{X}\}_{t = 1} ^ {T}} \sum_{t = 1} ^ {T}\big(\frac{\mu}{2} \norm{\x_{t} - \bx_t^\star} ^ {2} + \frac{1}{2} \norm{\x_{t} - \x_{t - 1}} ^ {2}\big) \label{constrained}, \\
			  & \stackrel{(c)}= \sum_{t = 1} ^ {T} f_{t}(\bx_t^\star) + \min_{\{\x_{t} \in \mathbb{R} ^ {d}\}_{t = 1} ^ {T}} \sum_{t = 1} ^ {T}\big(\frac{\mu}{2}\norm{\x_{t} - \bx_t^\star} ^ {2} + \frac{1}{2} \norm{\x_{t} - \x_{t - 1}} ^ {2}\big) \label{unconstrained}, \\
			  & \stackrel{(d)}= \sum_{t = 1} ^ {T} f_{t}(\bx_t^\star) + \min_{x_{t} ^ {(k)} \in \mathbb{R}}\sum_{t = 1} ^ {T} \sum_{k = 1} ^ {d} \bigg(\frac{\mu}{2}(x_{t} ^ {(k)} - x_{t} ^ {\star(k)}) ^ {2} + \frac{1}{2} (x_{t} ^ {(k)} - x_{t - 1} ^ {(k)}) ^ {2} \bigg), \nn\\
			  & \stackrel{(e)}= \sum_{t = 1} ^ {T} f_{t}(\bx_t^\star) + \min_{x_{t} ^ {(k)} \in \mathbb{R}} \sum_{k = 1} ^ {d}  \sum_{t = 1} ^ {T} \bigg(\frac{\mu}{2} (x_{t} ^ {(k)} - x_{t} ^ {\star(k)}) ^ {2} + \frac{1}{2} (x_{t} ^ {(k)} - x_{t - 1} ^ {(k)}) ^ {2}\bigg)\nn, \\
			  & \stackrel{(f)}= \sum_{t = 1} ^ {T} f_{t}(\bx_t^\star) + \min_{\{\tilde{\x}_{k} \in \mathbb{R} ^ {T}\}_{k = 1} ^ {d}}  \sum_{k = 1} ^ {d} \psi(\tilde{\x}_{k})\label{variable_change}, \\
			  &\stackrel{(g)}= \sum_{t = 1} ^ {T} f_{t}(\bx_t^\star) + \sum_{k = 1} ^ {d} \psi(\bar{\x}_{k}^\star)\label{to_continue},
			\end{align}
	where \((a)\) follows from the \(\mu\)-strong convexity of \(f_t\), \((b)\) follows from the optimality condition \(\ip{\nabla f_{t}(\bx_t^\star)}{\x - \bx_t^\star} \geq 0\), \(\forall\x \in \mathcal{X}\) of the minimization \(\bx_t^\star = \argmin_{\x \in \mathcal{X}} f_{t}(\x)\). To get \((c)\), we use Lemma \ref{constrained_unconstrained_equal} that the constrained optimization over \(\mathcal{X}\) is equivalent to an unconstrained optimization over \(\mathbb{R} ^{d}\). To get \((d)\), we have expressed the optimization over \({\x_{t} \in \mathbb{R} ^{d}}\) as an optimization over \(x_{t} ^ {(k)} \in \mathbb{R}\), where \(\x_{t} = [x_{t}^{(1)}, \ldots, x_{t} ^ {(d)}]\) and \(\x_{t}^{*} = [x_{t}^{\star(1)}, \ldots, x_{t}^{\star(d)}]\). To get \((e)\), we interchange the order of summation \(\sum_{t = 1} ^ {T}\sum_{k = 1}^{d}\to \sum_{k = 1} ^ {d}\sum_{t = 1}^{T}\), in \((f)\) we have transformed the unconstrained optimization in the \(\x_{t}\) space to \(\tilde{\x}_{k}\) space by defining \(\tilde{\x}_{k} = [x_{1} ^ {(k)}, \ldots, x_{T} ^ {(k)}] ^ {\T} \in \mathbb{R}^{T}\), \(\tilde{\bx}_{k}^\star = [x_{1} ^ {\star(k)}, \ldots, x_{T} ^ {\star(k)}]^{\T} \in \mathbb{R}^T\). It is important to note that while the original optimization was over \(\x_{t} \in \mathbb{R}^{d}\), the optimization in \eqref{variable_change} is over \(\tilde{\x}_{k} \in \mathbb{R}^{T}\), where \(T\) is the time horizon. Finally, we get \((g)\) by defining \(\bar{\x}_{k}^\star := \argmin_{\tilde{\x}_{k} \in \mathbb{R}^{T}} \psi(\tilde{\x}_{k})\), where the function \(\psi(\tilde{\x}_{k})\) is defined as 
	 \begin{align}\label{psi_def}
		\psi(\tilde{\x}_{k}) :=  \sum_{t = 1} ^ {T}\bigg(\frac{\mu}{2} (x_{t} ^ {(k)} - x_t ^ {\star(k)}) ^ {2} + \frac{1}{2} (x_{t} ^ {(k)} - x_{t - 1} ^ {(k)}) ^ {2}\bigg) = \frac{\mu}{2} (\tilde{\x}_{k} - \tilde{\bx}_{k}^\star) ^ {\T} (\tilde{\x}_{k} - \tilde{\bx}_{k}^\star) + \frac{1}{2} \tilde{\x}_{k} ^ {\T}\B \tilde{\x}_{k},
	\end{align}
	where the matrix \(\B\) of dimension \(T \times T\) is defined as \begin{align}\label{b_def}\B := \begin{bmatrix}
			2 & -1 & \cdots & \cdots & 0 \\
			-1 & 2 & -1 & \cdots & 0 \\
			\vdots  & \vdots  & \vdots & \vdots & \vdots  \\
			\cdots & \cdots & -1 & 2 & -1 \\
			0 & \cdots & \cdots & -1 & 1
		\end{bmatrix}.
		\end{align}
	Next, we consider the function \(\frac{1}{2}\tilde{\x}_{k} ^ {\T}\B \tilde{\x}_{k} = \sum_{t = 1} ^ {T} \frac{1}{2} (x_t ^ {(k)} - x_{t - 1} ^ {(k)}) ^ {2}\) whose Hessian equals to the matrix \(\B\) \eqref{b_def}. To proceed further, we first study the lower and upper bounds on \(\lambda_{i}(\B)\). For this we make use of the Gershgorin's Circle Theorem stated next. \begin{theorem}[\textit{Gershgorin's Circle Theorem}]\cite[Thm. 6.1.1]{horn2012matrix}\label{gersh}
		\textit{Let \(\A = [a_{ij}]\) be a \(n \times n\) matrix with complex entries, i.e., \(a_{ij} \in \C\), where \(\C\) denotes the space of complex numbers. Let \begin{align}\nn
				R_{i}(\A) = \sum_{j \neq i} \abs{a_{ij}}, \hspace{3mm} i = 1, \ldots, n,
			\end{align}
			denote the sum of absolute values of the off-diagonal entries in the \(i\)-th row of \(\A\), and consider the \(n\) Gershgorin discs\begin{align}\nn
				G_{i}(\A) = \{z \in \mathbf{C}: \abs{z - a_{ii}} \leq R_{i}(\A)\}, \hspace{3mm} i = 1, \ldots, n.
			\end{align}
		Then, the eigenvalues of \(\A\) belong to the union of the Gershgorin discs\begin{align}\nn
			G(\A) = \cup_{i = 1} ^ {n} G_{i}(\A),
		\end{align}}
	i.e., \(\lambda_{i}(\A) \in G(\A)\), \(i = 1, \dots, n\).
	\end{theorem}
	\begin{corollary}\label{eig:B}
		The eigenvalues of \(\B\) \eqref{b_def} lie in the interval \([0, 4]\), i.e., \(\lambda_{i}(\B) \in [0, 4]\) for all \(1 \leq i \leq T\).
	\end{corollary}
\begin{proof}
	 Since \(\B\) is symmetric, its eigenvalues are real. Therefore, it suffices to only consider the real part of the Gershgorin discs. Note that there are exactly \(3\) distinct Gershgorin discs corresponding to the matrix \(\B\). For \(i = 1\), the real part of \(G_{i}(\B) = \{z \in \C : \abs{z - 2} \leq 1\}\) corresponds to the interval \([1, 3]\). Similarly, for all $i$ such that \(2 \leq i \leq T - 1\), the real part of \(G_{i}(\B)\) corresponds to the interval \([0, 4]\). For \(i = T\), the real part of \(G_{i}(\B)\) corresponds to the interval \([0, 2]\). Thus, the real part of \(G(\B) = \cup_{i = 1}^{T}G_{i}(\B)\) corresponds to the interval \([0, 4]\). Applying Theorem \ref{gersh}, each eigenvalue \(\lambda_{i}(\B) \in [0, 4]\). This completes the proof. 
\end{proof}
	 Corollary \ref{eig:B} implies that \(\B \succeq \mathbf{0}\). Thus, the function $\frac{1}{2}\tilde{\x}_{k}^{\T}\B\tilde{\x}_{k} = \sum_{t = 1}^{T}\frac{1}{2}(x_{t}^{(k)} - x_{t - 1}^{(k)})$ is jointly convex in \(\tilde{\x}_{k}\). This implies that \(\psi(\tilde{\x}_{k})\) \eqref{psi_def} is strongly convex in \(\tilde{\x}_{k}\) since it is the sum of a strongly convex function \(\frac{1}{2}(\tilde{\x}_{k} - \tilde{\bx}_{k}^\star) ^ {\T} (\tilde{\x}_{k} - \tilde{\bx}_{k}^\star)\) and a convex function \(\frac{1}{2}\tilde{\x}_{k}^{\T}\B\tilde{\x}_{k}\). Thus, \(\psi(\tilde{\x}_{k})\) has a unique minimizer \(\bar{\x}_{k}^{\star}\), which is given by\begin{align}\label{unconstrained_minimizer}
		\bar{\x}_{k} ^\star = \argmin_{\tilde{\x}_{k}} \psi(\tilde{\x}_{k}) \implies \nabla \psi(\bar{\x}_{k}^{\star}) = \mathbf{0} \implies \mu(\bar{\x}_{k}^{\star} - \tilde{\x}_{k}^\star) + \B \bar{\x}_{k} ^\star = \mathbf{0} \implies \bar{\x}_{k} ^\star = \mu (\B + \mu \I) ^ {-1}\tilde{\x}_k^\star.
	\end{align}
	Define \(\A := \B + \mu\I\), thus \(\A \succ \mathbf{0}\) (\(\because \B \succeq \mathbf{0}\)). Since \(\A \succ \mathbf{0}\), \(\A^{-1}\) exists with \(\A^{-1} \succ \mathbf{0}\) (because \(\A\) and \(\A^{-1}\) have the same eigenvectors with \(\lambda_{i}(\A^{-1}) = \frac{1}{\lambda_{i}(\A)}\)). The optimal value \(\psi(\bar{\x}_k^\star)\) is obtained by evaluating \(\psi(\tilde{\x}_{k})\) at \(\tilde{\x}_{k} = \bar{\x}_{k}^\star\) \eqref{unconstrained_minimizer} and is given by \begin{align}
		\psi(\bar{\x}_{k} ^\star) &= \frac{\mu}{2} (\bar{\x}_{k} ^\star - \tilde{\x}_{k}^\star) ^ {\T} (\bar{\x}_{k} ^\star - \tilde{\x}_{k}^\star) + \frac{1}{2} \bar{\x}_{k} ^{\star\T}\B\bar{\x}_{k} ^\star,\nn \\
		&= \frac{\mu}{2} \norm{\mu \A ^ {-1} \tilde{\x}_{k}^\star - \tilde{\x}_{k}^\star} ^ {2} + \frac{\mu ^ 2}{2} (\A ^ {-1} \tilde{\x}_{k}^\star) ^ {\T} (\A - \mu\I) (\A ^ {-1} \tilde{\x}_{k}^\star),\nn \\
		&= \frac{\mu}{2} \bigg( \mu ^ 2\tilde{\x}_{k}^{\star\T}  \A ^ {-2} \tilde{\x}_{k}^\star + \tilde{\x}_{k} ^ {\star\T}\tilde{\x}_{k}^{\star} - 2\mu \tilde{\x}_{k} ^ {\star\T} \A ^ {-1} \tilde{\x}_{k}^\star\bigg) + \frac{\mu ^ 2}{2} \bigg(\tilde{\x}_{k} ^ {\star\T} \A ^ {-1} \tilde{\x}_k^\star - \mu\tilde{\x}_{k} ^ {\star\T} \A ^ {-2} \tilde{\x}_k^\star\bigg),\nn \\
		&= \frac{\mu}{2} \bigg(\tilde{\x}_k ^ {\star\T} \tilde{\x}_k^\star - \mu \tilde{\x}_k ^ {\star\T}\A ^ {-1} \tilde{\x}_k^\star \bigg) = \frac{\mu}{2} \tilde{\x}_{k} ^ {\star\T} (\I - \mu \A ^ {-1}) \tilde{\x}_{k}^\star \label{last_step}.
	\end{align}
	Next, we establish that, for the particular matrix \(\A = \B + \mu \I\) with \(\B\) defined as in \eqref{b_def}, the following holds: \begin{align}\nn
		\I - \mu \A ^ {-1} \succeq \frac{1}{\mu + 4} (\A - \mu\I).
	\end{align}
Note that this is equivalent to \begin{align}\label{another_proof}
	\bigg(1 + \frac{\mu}{\mu + 4}\bigg) \I \succeq \mu \A ^ {-1} + \frac{1}{\mu + 4} \A.
\end{align}
To establish \eqref{another_proof}, we first upper bound the eigenvalues of the matrix \(\mu \A^{-1} + \frac{1}{\mu + 4}\A\). The following Lemma establishes a relationship between the eigenvalues of \(\mu \A^{-1} + \frac{1}{\mu + 4}\A\) and the eigenvalues of \(\A\).\\
\begin{lemma}\label{rel_eig}\it{
	The eigenvalues of \((\mu \A^{-1} + \frac{1}{\mu + 4}\A)\) are related with the eigenvalues of \(\A\) via the following relationship:
	\begin{align}\label{eig_diff_mat}
		\lambda_{i}\bigg(\mu \A^{-1} + \frac{1}{\mu + 4}\A\bigg) = \frac{\mu}{\lambda_{i}(\A)} + \frac{1}{\mu + 4} \lambda_{i}(\A).
	\end{align}
	Further, the eigenvalues of \(\A\) are related with the eigenvalues of \(\B\) as \begin{align}\label{eig_A_B}
		\lambda_{i}(\A) = \lambda_{i}(\B) + \mu.
	\end{align}    
}
\end{lemma}
\begin{proof}
	Let \(\v_{i}\) be an eigenvector of \(\A\) corresponding to the eigenvalue \(\lambda_{i}(\A)\). Therefore, \(\A \v_{i} = \lambda_{i}(\A)\v_{i} \implies \frac{1}{\mu + 4} \A \v_{i} = \frac{1}{\mu + 4} \lambda_{i}(\A)\v_i\). Also note that \(\A^{-1}\v_{i} = \frac{1}{\lambda_{i}(\A)}\v_{i} \implies \mu \A^{-1}\v_{i} = \frac{\mu}{\lambda_{i}(\A)}\v_{i}\). Adding, we get \[\bigg(\mu \A^{-1} + \frac{1}{\mu + 4}\A\bigg)\v_{i} = \bigg(\frac{\mu}{\lambda_{i}(\A)} + \frac{1}{\mu + 4}\lambda_{i}(\A)\bigg)\v_{i}.\] Therefore, \(\v_{i}\) is an eigenvector of the matrix \(\mu \A^{-1} + \frac{1}{\mu + 4}\A\) with eigenvalue \(\frac{\mu}{\lambda_{i}(\A)} + \frac{1}{\mu + 4}\lambda_{i}(\A)\). 
	
	For the second part, let \(\u_{i}\) be an eigenvector of \(\B\) with eigenvalue \(\lambda_{i}(\B)\). Therefore, \(\B\u_{i} = \lambda_{i}(\B)\u_{i}\). Observe that \[(\B + \mu \I)\u_{i} = \lambda_{i}(\B) \u_{i} + \mu \u_{i} = (\lambda_{i}(\B) + \mu)\u_{i}.\] Hence, \(\u_{i}\) is an eigenvector of the matrix \(\B + \mu \I = \A\) with eigenvalue \(\lambda_{i}(\B) + \mu\). This completes the proof.
\end{proof}
From Corollary \ref{eig:B} and Lemma \ref{rel_eig} we have the following:
\begin{align}\label{applied_gershgorin}
	\lambda_{i}(\B) \in [0, 4] \stackrel{\eqref{eig_A_B}}\equiv \lambda_{i}(\A) - \mu \in [0, 4] \implies \lambda_{i} (\A) \in [\mu, \mu + 4].
\end{align}
Consider the function \(g(\lambda) = \frac{\mu}{\lambda} + \frac{\lambda}{\mu + 4}\) that is convex in \(\lambda\). The maximum value of this function over the interval \([\mu, \mu + 4]\) is attained at either of the end points \(\{\mu, \mu + 4\}\). Therefore, we have \begin{align} \label{max_bound}
	\max_{\lambda \in [\mu, \mu + 4]} g(\lambda) = \max\{g(\mu), g(\mu + 4)\}  = 1 + \frac{\mu}{\mu + 4}.
\end{align} Hence, from \eqref{eig_diff_mat}, \eqref{applied_gershgorin}, and \eqref{max_bound}, we obtain \begin{align}\label{eig_bound}\lambda_{i}\bigg(\mu\A ^ {-1} + \frac{1}{\mu + 4}\A\bigg) \leq 1 + \frac{\mu}{\mu + 4}, \end{align} for all \(1 \le i \le T\}\).
Since, \(\A, \A ^ {-1} \succ 0\), from the orthogonal decomposition theorem we have \begin{align}
	\mu \A^ {-1} + \frac{1}{\mu + 4} \A = \sum_{i = 1} ^ {T} \lambda_{i} \bigg(\mu \A ^ {-1} + \frac{1}{\mu + 4} \A\bigg) \v_{i} \v_{i} ^ {\T} \preceqtext{\eqref{eig_bound}} \sum_{i = 1} ^ {T} \bigg(1 + \frac{\mu}{\mu + 4}\bigg) \v_{i} \v_{i} ^ {\T} = \bigg(1 + \frac{\mu}{\mu + 4}\bigg)\I\nn,
\end{align}
where \(\v_{1}, \ldots, \v_{T}\) are the orthornomal eigenvectors corresponding to the \(T\) eigenvalues of \(\mu \A ^ {-1} + \frac{1}{\mu + 4} \A\). Thus, we have proved the claim in \eqref{another_proof}. Using \eqref{another_proof} in \eqref{last_step}, we obtain \begin{align}\label{imp_step}
\psi(\hat{\x}_{k} ^\star) \geq \frac{\mu}{2(\mu + 4)} \tilde{\x}_{k}^\star(\A - \mu\I) \tilde{\x}_k^\star = \frac{\mu}{2(\mu + 4)} \tilde{\x}_k ^ {\star\T} \B \tilde{\x}_k^\star = \frac{\mu}{2(\mu + 4)} \sum_{t = 1} ^ T (x_{t} ^ {\star(k)} - x_{t - 1} ^ {\star(k)} ) ^ {2},
\end{align}
where we have set \({\x}_0^\star = \mathbf{0}\). Continuing from step \eqref{to_continue}, we obtain \begin{align}
	C_\opt &\geq \sum_{t = 1} ^ {T} f_{t}(\bx^\star_t) + \sum_{k = 1} ^ {d} \psi(\bar{\x}_k ^\star),\nn \\
	&\geqtext{\eqref{imp_step}} \sum_{t = 1} ^ {T} f_t(\bx^\star_t) + \frac{\mu}{2(\mu + 4)} \sum_{k = 1}^{d}\sum_{t = 1}^{T}(x_{t} ^ {\star(k)} - x_{t - 1} ^ {\star(k)} ) ^ {2},\nn \\
	& = \sum_{t = 1} ^ {T} f_t(\bx^\star_t) + \frac{\mu}{2(\mu + 4)} \sum_{t = 1}^{T} \sum_{k = 1}^{d}(x_{t} ^ {\star(k)} - x_{t - 1} ^ {\star(k)} ) ^ {2},\nn \\
	&= \sum_{t = 1} ^ {T} f_t(\bx^\star_t) + \frac{\mu}{2(\mu + 4)} (\P_{2, T} ^\star + \norm{\bx^\star_{1}}^2)\nn.
\end{align}
This completes the proof.
\end{proof}
Next, we prove that the constrained optimization in \eqref{constrained} has the same solution as the unconstrained optimization in \eqref{unconstrained}. Although this has been shown in \cite{li2020online}, we provide a rigorous proof.
\begin{lemma}\label{constrained_unconstrained_equal}
	\textit{The constrained optimization in \eqref{constrained} has the same solution as the unconstrained optimization in \eqref{unconstrained}.}
\end{lemma}
\begin{proof}
	We first consider the special case of \(T = 2\) to build intuition. For this case, exact solutions to the unconstrained optimization in \eqref{unconstrained} are obtainable. Consider the unconstrained problem \eqref{unconstrained} with \(T = 2\), i.e., \begin{align}
		(\hat{\x}_{1}, \hat{\x}_2) := \argmin_{\x_{1}, \x_{2}} \bigg(\frac{\mu}{2} \norm{\x_{1} - \bx^\star_{1}} ^ {2} + \frac{\mu}{2} \norm{\x_{2} - \bx^\star_{2}} ^ {2} + \frac{1}{2} \norm{\x_{2} - \x_{1}} ^ {2} + \frac{1}{2}\norm{\x_{1}} ^ {2}\bigg)\nn.
	\end{align}
	The optimal solution can be expressed as the following system of equations: \begin{align}
		\mu(\hat{\x}_1 - \bx^\star_{1}) + (\hat{\x}_1 - \hat{\x}_2) + \hat{\x}_1 = \mathbf{0} & \implies (\mu + 2) \hat{\x}_{1}  - \hat{\x}_2 = \mu \bx^\star_{1}, \nn\\
		\mu(\hat{\x}_{2} - \bx^\star_{2}) + (\hat{\x}_{2}  - \hat{\x}_{1}) =  \mathbf{0} & \implies -\hat{\x}_{1}  + (\mu + 1) \hat{\x}_{2} = \mu \bx^\star_{2}.\nn
	\end{align}
	The optimal solution is given by \begin{align}\nn
		\begin{bmatrix}
			\hat{\x}_{1} \\ 
			\hat{\x}_{2} 
		\end{bmatrix} = \begin{bmatrix}
			\mu + 2 & -1 \\
			-1 & \mu + 1
		\end{bmatrix} ^ {-1} \begin{bmatrix}
			\mu \bx^\star_{1} \\
			\mu \bx^\star_{2}
		\end{bmatrix} = \frac{\mu}{\mu ^ {2} + 3\mu + 1} \begin{bmatrix} (\mu + 1)\x_1^{\star} + \x_2^\star \\
			\x_1^\star + (\mu + 2)\x_2^\star
		\end{bmatrix}.
	\end{align}
	Also, since \(\x_0 = \mathbf{0} \in \mathcal{X},\bx^\star_{1}, \bx^\star_{2} \in \mathcal{X}\), we can express \(\hat{\x}_{1}, \hat{\x}_{2} \) as a convex combination as \begin{align}
		& \hat{\x}_{1} = \bigg(\frac{\mu(\mu + 1)}{\mu ^ {2} + 3 \mu + 1} \bx^\star_{1} + \frac{\mu}{\mu ^ {2} + 3\mu + 1} \bx^\star_{2} + \frac{\mu + 1}{\mu ^ {2} + 3\mu + 1} \mathbf{0}\bigg) \in \mathcal{X}, \nn\\
		& \hat{\x}_{2} = \bigg( \frac{\mu}{\mu ^ {2} + 3\mu + 1} \bx^\star_{1} + \frac{\mu ^ {2} + 2\mu}{\mu ^ {2} + 3\mu + 1} \bx^\star_{2} + \frac{1}{\mu ^ {2} + 3\mu + 1} \mathbf{0}\bigg) \in \mathcal{X}. \nn
	\end{align}
	 From the considered special case, we observe that the optimal solution \(\hat{\x}_{1}, \hat{\x}_{2} \in \mathcal{X}\). Although, the coefficients corresponding to \(\x_{1}^\star, \x_{2}^\star\) in \(\hat{\x}_{1}\) are positive, they don't form a convex combination (\(\because\) they don't add up to \(1\)). But the remaining contribution can be attributed to \(\mathbf{0}\). This makes \(\hat{\x}_{1}\) a convex combination of \(\x_{1}^\star, \x_{2}^\star\), and \(\mathbf{0}\) each of which belongs to \(\mathcal{X}\). Similar reasoning can be applied to \(\hat{\x}_{2}\). Therefore, the unconstrained optimization is equivalent to the constrained optimization for \(T = 2\). Next, we proceed towards proving the Lemma for any general \(T > 2\).\\ 
	 Notice that the optimal solution \(\hat{\x} = [\hat{\x}_1, \ldots, \hat{\x}_{T}]\) of the unconstrained optimization in \eqref{unconstrained} can be expressed as the following system of equations (by setting the gradient = \(\mathbf{0}\)):\begin{align}
		\mu(\hat{\x}_{1} - \x_{1}^\star) + (\hat{\x}_{1} - \hat{\x}_{2}) + \hat{\x}_{1} = \mathbf{0} & \implies \bigg(1 + \frac{2}{\mu}\bigg)\hat{\x}_{1} - \frac{1}{\mu}\hat{\x}_{2} = \x_{1}^\star, \label{t_1} \\
		 \mu(\hat{\x}_{t} - \x_{t}^{\star}) + (\hat{\x}_{t} - \hat{\x}_{t - 1}) + (\hat{\x}_{t} - \hat{\x}_{t + 1}) = \mathbf{0} & \implies -\frac{1}{\mu}\hat{\x}_{t - 1} + \bigg(1 + \frac{2}{\mu}\bigg) \hat{\x}_{t} - \frac{1}{\mu}\hat{\x}_{t + 1} = \x_{t}^\star,\label{between_t} \\
		 \mu(\hat{\x}_{T} - \x_{T}^\star) + (\hat{\x}_{T} - \hat{\x}_{T - 1}) = \mathbf{0} &\implies -\frac{1}{\mu}\hat{\x}_{T - 1} + \bigg(1 + \frac{1}{\mu}\bigg)\hat{\x}_{T} = \x_{T}^{*}. \label{t_T}
	\end{align}
	Note that \eqref{between_t} holds  for all \(2 \leq t \leq T - 1\). Combining \eqref{t_1}, \eqref{between_t}, and \eqref{t_T} we can express the optimal solution \(\hat{\x}\) in the following matrix notation:
	 \begin{align}\label{matrix_eqn}
		\H \hat{\x} = \x^{*} \implies \begin{bmatrix}
			1 + \frac{2}{\mu} & -\frac{1}{\mu} & \ldots & \ldots & 0 \\
			-\frac{1}{\mu} & 1 + \frac{2}{\mu} & -\frac{1}{\mu} & \ldots & 0 \\
			\vdots & \vdots & \vdots & \vdots & \vdots \\
			\ldots & \ldots & -\frac{1}{\mu} & 1 + \frac{2}{\mu} & -\frac{1}{\mu} \\
			0 & \ldots & \ldots & -\frac{1}{\mu} & 1 + \frac{1}{\mu}
		\end{bmatrix}\begin{bmatrix}
			\hat{\x}_{1} \\ \vdots \\ \vdots \\ \hat{\x}_{T}
		\end{bmatrix} = \begin{bmatrix}
			\x_1^\star \\ \vdots \\ \vdots \\ \x_T^\star
		\end{bmatrix}.
	\end{align}
	Observe that the matrix \(\H\) has positive elements on it's main diagonal, negative elements = \(-\frac{1}{\mu}\) on  the diagonal above and below the main diagonal, and elements \(= 0\) everywhere else. Also, \(\H\) is strictly diagonally dominant (as per Definition \ref{diagonal_dominance_def}), therefore invertible.
	\begin{definition}\label{diagonal_dominance_def}
		\textit{A \(n \times n\) square matrix \(\X\) with elements \(x_{ij}\), is said to be diagonally dominant if, for every row of the matrix, the magnitude of the diagonal entry in a row is larger than or equal to the sum of the magnitudes of all the other off-diagonal entries in that row, i.e. \begin{align}\label{diagonally_dominant}\abs{x_{ii}} \geq \sum_{j \neq i} \abs{x_{ij}}, \forall i.
		\end{align}
	If a strict inequality is used in \eqref{diagonally_dominant} then \(\X\) is said to be strictly diagonally dominant.}
	\end{definition}
	 We now invoke \cite[Thm. 2]{varah1975lower} to prove a bound on the \(\infty\)-norm of the matrix \(\H\)\eqref{matrix_eqn}.
	\begin{theorem}\cite[Thm. 2]{varah1975lower}\label{varah}
		\textit{Assume a \(n \times n \) matrix \(\X\) with elements \(x_{ij}\), is diagonally dominant (refer to definition \eqref{diagonal_dominance_def}) and let \(\alpha_{\X} := \min_{k}(\abs{x_{kk}} - \sum_{j \neq k} \abs{x_{kj}})\). Then, \(\norm{\X ^ {-1}}_{\infty} < \frac{1}{\alpha_{\mathcal{\X}}}\), where \(\norm{\A}_{\infty} := \max_{1 \leq i \leq n} \sum_{j = 1} ^ {n} \abs{a_{ij}}\) denotes the \(\infty\)-norm of a \(n \times n\) matrix \(\A\) with elements \(a_{ij}\).}
	\end{theorem}For the matrix \(\H = [h]_{ij}\) \eqref{matrix_eqn}, we have \(
		\alpha_{\H} = \min_{k}(\abs{h_{kk}} - \sum_{j \neq k}\abs{h_{kj}}) = 1 \). Also \(\H\) is strictly diagonally dominant. Thus, applying Theorem \ref{varah}, we obtain \begin{align}\label{sum_norm}
			\norm{\H^{-1}}_{\infty} < 1.
		\end{align}
	Note that the entries of \(\H\) are not positive. However, we show that all the entries of \(\H^{-1} = [\tilde{h}]_{ij}\) are positive. The special structure of \(\H\) has been studied in \cite{concus1985block} and exact analytical expressions for \(\tilde{h}_{ij}\) can be found in \cite{li2020online}. However \cite{li2020online} do not provide a rigorous proof for \(\tilde{h}_{ij} > 0\). We prove this in the following Lemma. 
	\begin{lemma}\label{positive}
		\textit{The elements of  \(\H^{-1} = [\tilde{h}_{ij}]\) satisfy \(\tilde{h}_{ij} > 0\), for all \(1 \leq i \leq T\) and \(1 \leq j \leq T\).}
	\end{lemma}
	\begin{proof}
		Recall from \eqref{matrix_eqn} that each entry of \(\H\) is a function of \(\mu\). From \cite{li2020online}, the elements \(\tilde{h}_{t, t + \tau}\) where \(\tau \geq 0\) have a closed form expression given by \(\tilde{h}_{t, t + \tau} = \mu w_{t} v_{t + \tau}\), where \begin{align}
			w_{t} = \frac{\rho}{1 - \rho ^ {2}}\bigg(\frac{1}{\rho ^ {t}} - \rho ^ {t}\bigg), v_{t} = c_{1} \frac{1}{\rho ^ {T - t}} + c_{2} \rho ^ {T - t}, v_{T} = \frac{1}{(\xi - 1) w_{T} - w_{T - 1}}, \label{analytical_expression} \\
			c_{1} := v_{T}\bigg(\frac{(\xi - 1)\rho - \rho ^ {2}}{1 - \rho ^ {2}}\bigg), c_{2} := v_{T} \frac{1 - (\xi - 1)\rho}{1 - \rho ^ {2}}\nn,
		\end{align}
	and \(\rho := \frac{\sqrt{\mu + 4} - \sqrt{\mu}}{\sqrt{\mu + 4} + \sqrt{\mu}}, \xi := \mu + 2\). Note that since \(\H\) is symmetric, \(\H ^ {-1}\) is symmetric as well. Therefore, to prove that all entries \(\tilde{h}_{ij} > 0\), it is sufficient to consider the elements \(\tilde{h}_{t, t + \tau}\) where \(\tau \geq 0\). The collection of elements \(\{\tilde{h}_{t, t + \tau} : 1 \leq t \leq T, 0\leq \tau \leq T - \tau\}\), corresponds to the elements in the upper triangular block of \(\H^{-1}\). Note that \(\rho < 1\), and thus \(w_{t} > 0\), for all \(1 \leq t \leq T\).

	To prove the Lemma, it is sufficient to show that \(v_{t} > 0\), for all \(1 \leq t \leq T\). To prove this, we first show that \(v_{T} > 0\). From \eqref{analytical_expression}, \(v_{T}\) depends on \(w_{T}\) and \(w_{T - 1}\) which are given by\begin{align}
		w_{T} = \frac{\rho}{1 - \rho ^ {2}} \bigg(\frac{1}{\rho ^ {T}} - \rho ^ {T}\bigg),w_{T - 1} = \frac{\rho}{1 - \rho ^ {2}} \bigg(\frac{1}{\rho^{T - 1}} - \rho ^ {T - 1}\bigg).
	\end{align}

	Consider the function \(\zeta(x) = \frac{1}{x} - x\) that is monotonically decreasing in \(x\), since \(\zeta'(x) = -\frac{1}{x ^ {2}} - 1 < 0\). Since \(\rho ^ {t}\) decreases as \(t\) increases (\(\because \rho < 1\)), the quantity $\frac{1}{\rho ^ {t}} - \rho ^ {t}$ increases as \(t\) increases. Therefore, \(w_{T} > w_{T - 1}\). This implies that \(v_{T} = \frac{1}{(\xi - 1)w_{T} - w_{T - 1}} > 0\), since \(\xi > 2\). Next we show that \(v_{t} > 0\) for all \(1 \leq t \leq T - 1\).

	Since \(v_{T} > 0\) and \(\rho < 1\), we have \(c_{1} = v_{T} \rho \big(\frac{(\mu + 1) - \rho}{1 - \rho ^ {2}}\big) > 0\). Note that, we have substituted \(\xi = \mu + 2\) to obtain \(c_{1}\). Similarly, \(c_{2} = v_{T} \frac{1 - (\mu + 1)\rho}{1 - \rho ^ {2}}\). To prove that \(c_{2} > 0\), it is sufficient to prove that the quantity \(1 - (\mu + 1)\rho > 0\). Note that \(1 - (\mu + 1)\rho = 1 - (\mu + 1)\frac{\sqrt{\mu + 4} - \sqrt{\mu}}{\sqrt{\mu + 4} + \sqrt{\mu}} > 0,\) for all \(\mu > 0\). Therefore, \(c_{2} > 0\).

	Since \(c_{1} > 0\) and \(c_{2} > 0\), we have \(v_{t} > 0\), for all \(1 \leq t \leq T - 1\). 
	With \(v_{t} > 0, w_{t} > 0\), for all \(1 \leq t \leq T\), we have \(\tilde{h}_{t, t+\tau} > 0\). This completes the proof.
	\end{proof}
From \eqref{matrix_eqn}, the optimal solution \(\hat{\x}_{k}\) can be expressed as \begin{align}\nn
		\hat{\x}_{k} = \sum_{j = 1} ^ {T} \tilde{h}_{kj} \bx^\star_{j}.
	\end{align}
	Observe from \eqref{sum_norm} that for every \(k\), \(\sum_{j = 1} ^{T} \abs{\tilde{h}_{kj}} < 1\). Also, since \(\tilde{h}_{kj} > 0\) (from Lemma \ref{positive}), we can express \(\hat{\x}_{k}\) as \begin{align}
		\hat{\x}_{k}  = \sum_{j = 1} ^ {T} \tilde{h}_{kj} \bx^\star_{j} + \big(1 - \sum_{j = 1} ^ {T} \tilde{h}_{kj}\big) \mathbf{0}. \nn
	\end{align}
	This clearly implies \(\tilde{\x}_{k} \in \mathcal{X}\), for all \(k\), since \(\x_{t}^{\star} \in \mathcal{X},\) for all  \(1 \leq t \leq T\), \(\mathbf{0} \in \mathcal{X}\), and \(\mathcal{X}\) is convex. This completes the proof.
\end{proof}

\begin{lemma}\label{f_tsame}
\textit{If the functions \(f_{t}\) are identically equal to a function \(f\), i.e., \(f_{t} = f\), for all \( 1 \leq t \leq T\), the cost incurred by the $\opt$ can be lower bounded by \begin{align}
	C_\opt \geq Tf(\x^\star) +  \frac{\mu}{2(\mu + 1)} \norm{\x^\star} ^ {2}\nn,
\end{align}
where \(\x^\star:=\argmin_{\x \in \mathcal{X}} f(\x)\).}
\end{lemma}
\begin{proof}
 Continuing from step \eqref{unconstrained} as in the proof of Lemma \eqref{lem:lbOPT}, we have the following:		
	\begin{align}
		C_\opt \geq T f(\x^\star) + \min_{\{\x_{t} \in \mathbb{R} ^ {d}\}_{t = 1} ^ {T}} \sum_{t = 1} ^ {T} \big(\frac{\mu}{2}\norm{\x_{t} - \x^\star} ^ {2} + \frac{1}{2} \norm{\x_{t} - \x_{t - 1}} ^ {2} \big),\nn
	\end{align}
Define \(C_\opt(j)\) to be the following:
\begin{align}
	C_\opt(j) := Tf(\x^\star) + \min_{\{\x_{t} \in \mathbb{R} ^ {d}\}_{t = 1} ^ {j}} \sum_{t = 1} ^ {j} \big(\frac{\mu}{2}\norm{\x_{t} - \x^\star} ^ {2} + \frac{1}{2} \norm{\x_{t} - \x_{t - 1}} ^ {2} \big),\nn
\end{align}
It is trivial to notice that \(\{C_{\opt}(j)\}_{j = 1} ^ {T}\) is a monotonically increasing sequence. To obtain a lower bound on \(C_\opt\), we can use \(C_\opt(1)\) and obtain the following:
\begin{align}\label{first_bound}
	C_\opt \geq C_\opt(T) \geq C_{\opt}(1) = Tf(\x^\star) + \min_{\x_{1} \in \mathbb{R} ^ {d}} \big(\frac{\mu}{2} \norm{\x_{1} - \x^\star} ^ {2} + \frac{1}{2} \norm{\x_{1}} ^ {2} \big),
\end{align}
The unconstrained minimizer of $\frac{\mu}{2} \norm{\x_{1} - \x^\star} ^ {2} + \frac{1}{2} \norm{\x_{1}} ^ {2} $ is given by
\begin{align}
	\argmin_{\x_{1} \in \mathbb{R} ^ {d}}\big(\frac{\mu}{2}\norm{\x_{1} - \x^\star} ^ {2} + \frac{1}{2} \norm{\x_{1}} ^ {2} \big) = \frac{\mu}{\mu + 1} \x^\star\nn.
\end{align}
Therefore, \eqref{first_bound} reduces to
\begin{align}
	C_\opt \geq Tf(\x^\star) + \frac{\mu}{2} \|{\frac{\mu}{\mu + 1} \x^\star - \x^\star}\| ^ {2} + \frac{1}{2} \|{\frac{\mu}{\mu + 1}\x^\star}\| ^ {2} = Tf(\x^\star) + \frac{\mu}{2(\mu + 1)} \norm{\x^\star} ^ {2}.\nn
\end{align}
This completes the proof.
\end{proof}
Note that the result of Lemma \ref{f_tsame} is similar to the bound in Lemma \ref{lem:lbOPT} and implies that as \(\mu \to 0\), the cost incurred by the $\opt$ is lower bounded by \(\Omega(Tf(\x^\star) + \mu \norm{\x^\star} ^ {2})\).

\section{Upper bounding \(C_{\cA_o}\) when \(\sum_{t = 1} ^ {T} \norm{f_{t}(\x_{t}^\star)} ^ {2} = \mathcal{O}(\cP_{2, T}^\star)\)}\label{app:remark_6}
We show that as long as the quantity \(\sum_{t = 1}^{T}\norm{\nabla f_t(\x_{t}^\star)}^{2} = \mathcal{O}(\mathcal{P}_{2, T}^{*})\), we get the same order-wise (with respect to \(L, \mu\)) bound on the competitive ratio as Theorem \ref{thm:cromgd}. Let \(\sum_{t = 1}^{T} \norm{\nabla f_{t}(\x_{t}^\star)}^{2} = \gamma \mathcal{P}_{2, T}^\star\) for some constant \(\gamma > 0\) (that is independent of \(T\) and other problem parameters, i.e., \(\mu, L\), etc.). Then, the upper bound on \(C_{\cA_o}\) in Lemma \ref{lem:upOMGD} is a function of \(\alpha\), which can therefore be minimized with respect to \(\alpha\) as done in the following steps: \begin{align}
	C_{\mathcal{A}_o} &\leq \sum_{t = 1} ^ {T} f_{t}(\x_{t}^\star) + \frac{\gamma}{2\alpha}\mathcal{P}_{2, T}^\star + \alpha (\norm{\x_{1} - \x_{1}^\star}^{2} + 2 \mathcal{P}_{2, T}^\star)+ (L + 5) (\norm{\x_{1} - \x_{1}^\star}^{2} + 2\mathcal{P}_{2, T}^{*}), \nn \\
	& \leq \sum_{t = 1} ^ {T} f_{t}(\x_{t}^\star) + \big(\frac{\gamma}{2\alpha} + 2\alpha + 2L + 10\big) \big(\norm{\x_{1} - \x_{1}^\star} ^ {2} + \mathcal{P}_{2, T}^\star), \label{minimize_alpha} \\
	& \stackrel{(a)}\leq \sum_{t = 1} ^ {T} f_{t}(\x_{t}^\star)  + (2L + 2\sqrt{\gamma} + 10) (\norm{\x_{1} - \x_{1}^\star}^{2} + \mathcal{P}_{2, T}^{*}) \nn,
\end{align}
where \((a)\) follows by minimizing the upper bound in \eqref{minimize_alpha} with respect to \(\alpha\), the optimal choice of \(\alpha\) being \(\argmin_{\alpha > 0} \big(\frac{\gamma}{2\alpha} + 2\alpha \big) = \frac{\sqrt{\gamma}}{2}\).
Next, proceeding similarly as the proof of Theorem \ref{thm:cromgd}, the competitive ratio of the OMGD algorithm is bounded by \(
\frac{C_{\mathcal{A}_o}}{C_{\opt}} \leq 4(L + \sqrt{\gamma} + 5) + \frac{16(L + \sqrt{\gamma} + 5)}{\mu},
\)
which is order-wise (with respect to \(L, \mu\)) same as the result of Theorem \ref{thm:cromgd}.
\section{Proof of Lemma \ref{lb:omgd}}\label{app:lb:omgd}
We consider \(\cX = \Rn ^ d\). Let \(T = 2\) and consider \(f_{1}(\x) = \frac{\mu}{2}\norm{\x - \th} ^ {2}\), where \(\th \ne \mathbf{0}\), and \(f_{2}(\x) = \frac{L}{2}\norm{\x}^{2}\), where \(L \geq \mu\). Note that both $f_1$ and $f_2$ and $L$-smooth and $\mu$-strongly convex. Consider the starting point \(\x_{0} = \mathbf{0}\). For this choice of functions, \(\cA_o\) selects \(\x_{1} = \x_{0} = \mathbf{0}\). However, it takes \(K = \lceil\frac{L + \mu}{2\mu}\ln 4 \rceil\) gradient steps with step size \(\frac{1}{L}\), starting at \(\z_{2} ^ {(0)} = \mathbf{0}\). The progress from \(\z_{2} ^ {(k - 1)}\) to \(\z_{2} ^ {(k)}\) happens through the following update: \begin{align}\nn
		\z_{2}^{(k)} = \z_{2} ^ {(k - 1)} - \frac{\mu}{L}(\z_{2} ^ {(k - 1)} - \th) = \z_{2} ^{(k - 1)}\big(1 - \frac{\mu}{L}\big) + \frac{\mu}{L}\th. \end{align}
	The final action \(\z_{2} ^ {(K)}\) is obtained by unrolling the recursion and is given by
	\begin{align}
		\z_{2} ^ {(K)} &=  \z_{2} ^{(K - 1)}\big(1 - \frac{\mu}{L}\big) + \frac{\mu}{L}\th, \nn \\
		&= \z_{2}^{(K - 2)}\big(1 - \frac{\mu}{L}\big) ^ {2} + \frac{\mu}{L}\th \big(1 + \big(1 - \frac{\mu}{L}\big)\big), \nn\\
		&= \z_{2} ^ {(K - 3)}\big(1 - \frac{\mu}{L}\big) ^ {3} + \frac{\mu}{L}\th\big(1 + \big(1 - \frac{\mu}{L}\big) + \big(1 - \frac{\mu}{L}\big)^{2} \big), \nn\\
		& \hspace{3mm}\vdots\hspace{15mm}\vdots \hspace{15mm} \vdots \hspace{15mm}, \vdots \hspace{15mm} \vdots \nn\\
		&= \z_{2} ^ {(0)}\big(1 - \frac{\mu}{L}\big) ^ {K} + \frac{\mu}{L}\th \sum_{i = 0} ^{K - 1} \big(1 - \frac{\mu}{L}\big) ^ {i}, \nn\\
		&= \z_{2} ^ {(0)}\big(1 - \frac{\mu}{L}\big) ^ {K} + \th\big(1 - \big(1 - \frac{\mu}{L}\big) ^ {K}\big), \nn \\
		&=  \th\big(1 - \big(1 - \frac{\mu}{L}\big) ^ {K}\big).
	\end{align}
	Since \(\mathcal{A}_{o}\) selects \(\x_{2} = \z_{2}^{(K)}\), the resulting cost incurred by \(\mathcal{A}_{o}\) is  \[
	C(\mathcal{A}_{o}) = \frac{\mu}{2}\norm{\th} ^ {2} + \frac{L + 1}{2}\norm{\x_{2}} ^ {2} = \frac{\mu}{2}\norm{\th} ^ {2} + \frac{L + 1}{2}\big(1 - \big(1 - \frac{\mu}{L}\big) ^ {K}\big) ^ {2}\norm{\th} ^ {2}.\]
	Since \(e ^{x} \geq 1 + x\), for all \(x \in \mathbb{R}\), we have \(e ^ {-\frac{\mu}{L}}\geq \big(1 - \frac{\mu}{L}\big) \implies 1 - \big(1 - \frac{\mu}{L}\big) ^ {K} \geq 1 - e ^ {-K\frac{\mu}{L}}\). The quantity \(1 - e ^ {-K\frac{\mu}{L}}\) can be lower bounded in the following way:
	\begin{align}\nn
		1 - e ^ {-K\frac{\mu}{L}} = 1 - e ^ {-\ceil{\frac{L + \mu}{2\mu}\ln 4}\frac{\mu}{L}} \geq 1 - e ^ {-\frac{L + \mu}{2\mu}\frac{\mu}{L}\ln 4} = 1 - e ^ {-\frac{L + \mu}{L} \ln 2} = 1 - \frac{1}{2 ^ {1 + \frac{\mu}{L}}} \geq \frac{1}{2}. 
	\end{align}
	Therefore, \(C_{\cA_o}\) can be lower bounded by
	\[
	C_{\mathcal{A}_{o}} \geq \frac{\mu}{2}\norm{\th}^{2} + \frac{L + 1}{8} \norm{\th}^{2}.
	\]
	The \(\opt\) on the other hand incurs the following cost:
	\begin{align}
		C_{\opt} &= \min_{\x_{1}, \x_{2}} \big(\frac{\mu}{2}\norm{\x_{1} - \th} ^ {2} + \frac{L}{2}\norm{\x_{2}}^{2} + \frac{1}{2}\norm{\x_{1}} ^ {2} + \frac{1}{2}\norm{\x_{1} - \x_{2}} ^ {2}\big)\nn, \\
		&\stackrel{(a)}\leq \min_{\x_{1}}\big(\frac{\mu}{2}\norm{\x_{1} - \th}^{2} + \norm{\x_{1}} ^ {2}\big) = \frac{\mu}{\mu + 2} \norm{\th} ^ {2}\nn,
	\end{align}
	where \((a)\) follows by setting \(\x_{2} = \mathbf{0}\) in the \(\opt\)'s minimization problem. Therefore, the competitive ratio of the OMGD algorithm \(\cA_o\) is lower bounded by \[
	\frac{C_{\mathcal{A}_{o}}}{C_\opt} \geq \frac{(4\mu + L + 1)(\mu + 2)}{8\mu}.
	\]
	Note that \((4\mu + L + 1)(\mu + 2) = 4\mu ^ {2} + 8\mu + (L + 1)(\mu + 2) > 8\mu + 2(L + 1) + \mu(L + 1)\). Hence, $\frac{C_{\cA_o}}{C_\opt}$ can be lower bounded by \(
	\frac{C_{\mathcal{A}_{o}}}{C_\opt} \geq 1 + \frac{L + 1}{4\mu} + \frac{L + 1}{8}.
	\)
	This completes the proof.
\section{Proof of Lemma \ref{lb:slowalg}}\label{app:lbslow}
	Sufficient to consider \(d = 1\) (one-dimensional case) and \(\cX = \Rn\). Consider the case when an algorithm $\cA$ is slow and $\epsilon_t = \cO(\mu)$ and $\mu\rightarrow 0$.
	Let $f_t(x) = (x-1)^2$, for all $t=1, \dots, T$. On account of $\cA$ being slow, the earliest time $t$ such that $\cA$'s action $x_t\ge \frac{1}{2}$ is $\Omega(\frac{1}{\mu})$. Thus, the cost of $\cA$ is at least $\Omega(\frac{1}{\mu})$. An algorithm $\cB$ that chooses $x_t=1$, for all $t$ has the cost of $1$. Thus, $C_\opt \le 1$, and consequently the competitive ratio of $\cA$ is at least 
	$\Omega(\frac{1}{\mu})$.

	Next, consider the case when an algorithm $\cA$ is slow and $\epsilon_t = \cO(\frac{1}{L})$ and $L\rightarrow \infty$.
	Let $f_t(x) = \frac{L}{2}(x-1)^2$, for all $t=1, \dots, T$. On account of $\cA$ being slow, the earliest time $t$ such $\cA$'s action $x_t\ge \frac{1}{2}$ is $\Omega(L)$. Thus the cost of $\cA$ is at least $\Omega(L^2)$. An algorithm $\cB$ that chooses $x_t=1$, for all $t$ has the cost of $1$. Thus, $C_\opt \le 1$. Therefore the competitive ratio of $\cA$ is at least 
	$\Omega(L^2)$. Combining both the cases completes the proof.
\section{Proof of Lemma \ref{preliminary_lower_bound}}\label{app:lb1}
	Sufficient to consider \(d = 1\) (one-dimensional case) and \(\cX = \Rn\). With the limited information setting, any online algorithm $\cA$ chooses $x_1=x_0={0}$, since otherwise $f_1(x)= \frac{L}{2}x^2$ and $\cA$'s competitive ratio is unbounded. Let $f_1(x) = \frac{L}{2}(x-1)^2$. Then the cost of $\cA$ is $\frac{L}{2}$. $\opt$ on the other hand chooses $x_1=1$ and pays only the switching cost of $\frac{1}{2}$. Thus, the competitive ratio of $\cA$ is at least $L$. This completes the proof.
\section{Proof of Lemma \ref{modified_preliminary_lower_bound}}\label{app:lb2}
	Sufficient to consider $d=1$ (one-dimensional case) and \(\cX = \Rn\).
	Choose $f_t(x)=\frac{\mu}{2}x^2$, for all $t=1,\dots, T'$. Any online algorithm $\cA$ has to choose $x_t=0$, for all $t=1,\dots, T'$ since otherwise its cost is non-zero while that of $\opt$ is $0$. Let $f_{T'+1}(x) = \frac{L}{2}(x - 1)^2$. Since $\cA$ chooses its action $x_{T'+1}$ before any information about $f_{T'+1}$ is revealed, thus, $x_{T'+1}=0$ as well and $\cA$'s cost is at least $\frac{L}{2}$ . The $\opt$, however, knowing the full sequence $\{f_t\}_{t = 1}^{T'+1}$, moves its actions slowly towards $1$ as a function of $\mu$, and as shown in \cite{goel2019beyond} for this choice of functions, $\opt$'s cost is $\mathcal{O}(\sqrt{\mu})$ as $T'\rightarrow \infty$ and \(\mu \to 0^{+}\). Thus, letting $T'\rightarrow \infty$ the competitive ratio of $\cA$ is at least $\Omega\left(\frac{L}{\sqrt{\mu}}\right)$ as \(\mu \to 0\). This completes the proof.
\section{Proof of Theorem \ref{regret_omgd}}\label{app:regret_thm}
	We first bound the quantity \(\sum_{t = 1}^{T} \norm{\x_{t} - \x_{t}^\star}\) for \(\mathcal{A}_o\) as follows.
	\begin{align}
		\sum_{t = 2} ^ {T} \norm{\x_{t} - \x_{t}^\star} & \stackrel{(a)}\leq \sum_{t = 2} ^ {T} \big(\norm{\x_{t} - \x_{t - 1}^\star} + \norm{\x_{t}^\star - \x_{t - 1}^\star}\big), \nn\\
		& \stackrel{(b)}\leq \sum_{t = 2} ^ {T} \frac{1}{2}\norm{\x_{t - 1} - \x_{t - 1}^\star} + \mathcal{P}_{T}^{\star}, \nn\\
		& \stackrel{(c)}= \frac{1}{2}\norm{\x_{1}^\star} + \sum_{t = 2} ^ {T - 1} \frac{1}{2}\norm{\x_{t} - \x_{t}^\star} + \mathcal{P}_{T}^\star, \nn \\
		& \leq \frac{1}{2}\norm{\x_{1}^\star} + \sum_{t = 2} ^ {T} \frac{1}{2}\norm{\x_{t} - \x_{t}^\star} + \mathcal{P}_{T}^\star, \nn,
	\end{align}
	where \((a)\) follows from the triangle inequality, \((b)\) follows from the contraction bound \eqref{one_fourth_0} attained by \(\mathcal{A}_{o}\), and \((c)\) follows since \(\cA_o\) chooses \(\x_{1} = \x_{0} = \mathbf{0}\). Rearranging, we obtain \begin{align}
		\sum_{t = 2} ^ {T} \norm{\x_{t} - \x_{t}^\star} \leq \norm{\x_{1}^\star} + 2\mathcal{P}_{T}^\star \implies \sum_{t = 1} ^ {T} \norm{\x_{t} - \x_{t}^\star} \leq 2\norm{\x_{1}^\star} + 2\mathcal{P}_{T}^\star. \label{cumulative_sum}
	\end{align}
	From the convexity of \(f\), we have the following:
	\begin{align}
		f_{t}(\x_{t}) - f_{t}(\x_{t}^\star) \leq \ip{\nabla f(\x_{t})}{\x_{t} - \x_{t}^\star} \stackrel{(a)}\leq \norm{\nabla f_{t}(\x_{t})} \norm{\x_{t} - \x_{t}^\star} \stackrel{(b)}\leq G \norm{\x_{t} - \x_{t}^\star},\nn
	\end{align} 
	where \((a)\) follows from the Cauchy-Schwarz inequality, while \((b)\) follows from assumption \ref{bounded_gradients}. Summing over \(t = 1, \ldots, T\), we obtain the following:
	\begin{align}
		& \sum_{t = 1} ^ {T} f_{t}(\x_{t}) - f_{t}(\x_{t}^\star) \leq G \sum_{t = 1} ^ {T}\norm{\x_{t} - \x_{t}^\star} \stackrel{\eqref{cumulative_sum}}\leq 2G(\norm{\x_{1}^\star} + \mathcal{P}_{T}^\star), \nn \\
		& \sum_{t = 1} ^ {T} f_{t}(\x_{t}) + \frac{1}{2}\norm{\x_{t} - \x_{t - 1}}^{2} \stackrel{(a)}\leq \sum_{t = 1} ^ {T} f_{t}(\x_{t}^\star) + 2G(\norm{\x_{1}^\star} + \mathcal{P}_{T}^\star) + 5 \norm{\x_{1}^\star}^{2} + 10 \mathcal{P}_{2, T}^\star \label{cost_bounded_subgradient},
	\end{align}
	where \((a)\) follows from the bound on the total switching cost attained by \(\mathcal{A}_{o}\) as given by Lemma \ref{lemma_switching_cost}.
	Finally, we can bound the regret as follows.\begin{align}
		R_d(\mathcal{A}_o) = C_{\mathcal{A}_o} - C_\opt \leqtext{\eqref{actual_lower_bound}, \eqref{cost_bounded_subgradient}}  2G( \norm{\x_{1}^\star} + \mathcal{P}_{T}^{*}) + 5 \norm{\x_{1}^{*}}^{2} + \bigg(10 - \frac{\mu}{2(\mu + 4)}\bigg) \mathcal{P}_{2, T}^\star.\nn
	\end{align}
	This completes the proof.
\section{Proof of Corollary \ref{cor:regert_omgd}}\label{app:cor_regret}
	When \(\mathcal{X}\) has a finite diameter, we can bound the quantity \(\alpha_{1} \norm{\x_{1}^\star}^{2} + \alpha_{2} \mathcal{P}_{2, T}^{*}\), where \(\alpha_{1}, \alpha_{2} > 0\) by the following:\begin{align}\label{triv_ineq}
		\alpha_{1} \norm{\x_{1}^\star}^{2} + \alpha_{2} \mathcal{P}_{2, T}^{*} \stackrel{(a)}\leq \max(\alpha_{1}, \alpha_{2})D (\norm{\x_{1}^\star} + \mathcal{P}_{T}^\star),
	\end{align}
	where \((a)\) holds because \(\alpha_{1}\norm{\x_{1}^\star}^{2} \leq \max(\alpha_{1}, \alpha_{2}) D\norm{\x_{1}^\star}\), and each of the terms \(\norm{\x_{t}^\star - \x_{t - 1}^\star}^{2}\) in \(\mathcal{P}_{2, T}^\star\) can be bounded by \(D\norm{\x_{t}^\star - \x_{t - 1}^\star}\).
	Therefore, we can bound \( R_d(\mathcal{A}_o)\) for the OMGD algorithm by the following:\begin{align}
		R_d(\mathcal{A}_{o}) &\stackrel{(a)}\leq \bigg(2G + D\bigg( 10 - \frac{\mu}{2(\mu + 4)}\bigg)\bigg)(\norm{\x_{1}^\star} + \mathcal{P}_T^\star),
	\end{align}
	where \((a)\) follows by applying \eqref{triv_ineq} to the regret bound on \(R_{d}(\cA_o)\) in Theorem \ref{regret_omgd}. This completes the proof.
\section{Discussion regarding bound \eqref{vt_sq_regret}}\label{app:vt_sq_regret}
We begin by proving the bound \eqref{vt_sq_regret} in the following manner:
\begin{align} \nn
	R_d(\mathcal{A}_o) & \stackrel{(a)}\leq\sum_{t = 1} ^ {T} f_{t}(\bx_t^\star) + (L + 5) (\norm{\bx^\star_{1}} ^ {2} + 2\P_{2, T} ^\star) - \sum_{t = 1} ^ {T} f_{t}(\bx_t^\star) - \frac{\mu}{2\mu + 8}(\P_{2, T} ^\star + \norm{\bx^\star_{1}} ^ {2}), \\
	&= \big(2L + 10 - \frac{\mu}{2\mu + 8}\big) (\P_{2, T} ^\star + \norm{\bx^\star_{1}} ^ {2})\label{tight_step},  \\
	&  \stackrel{(b)}\leq \big(2L + 10 - \frac{\mu}{2\mu + 8}\big)  \big(\sum_{t = 2} ^ {T} \norm{\bx_t^\star - \bx^\star_{t - 1}} + \norm{\x_{1}^\star}\big) ^ {2} \label{loose_step},  \\
	& \leq \big(2L + 10 - \frac{\mu}{2\mu + 8}\big) V_{T} ^ {2}, \nn
\end{align}
where \((a)\) follows from the bounds on \(C_{\mathcal{A}_o}\) and \(C_\opt\) in Theorem \ref{thm:cromgd}, while \((b)\) follows by the trivial inequality \(\sum_{i = 1} ^ {T} a_{i} ^ {2} \leq \big(\sum_{i = 1} ^ {T} a_{i}\big) ^ {2},\) for all  \(a_{1}, \ldots, a_{T} \geq 0\). Intuitively, when \(\mathcal{P}_{T}^\star + \norm{\x_{1}^\star}\) is sufficiently large (which is the case when the optimizers \(\x_{t}^\star\) and \(\x_{t - 1}^\star\) are far apart), moving from step \eqref{tight_step} to \eqref{loose_step} adds in a significant number of extra terms. Therefore, \eqref{loose_step} is meaningful only in the 
$V_{T} \to 0^{+}$ regime. 
\section{Proof of Lemma \ref{lem:lb_linear}}\label{app:lb_linear}
Sufficient to consider \(d = 1\) (one-dimensional case) and \(\cX = \Rn\). With the limited information setting, any online algorithm \(\cA\) has to choose \(x_{1} = x_{0} = 0\), since otherwise the adversary can choose \(f_{1}(x) = \frac{L}{2}x ^ 2\) and \(\cA\)'s competitive ratio is unbounded. 

Choose \(f_{1}(x) = \frac{\mu}{2}(x - \theta) ^ 2\), where \(\frac{\sqrt{129} - 9}{2\mu} \ge \theta > \frac{1}{\mu}\) and \(f_{2}(x) = \frac{\mu}{2}x ^ 2\). As argued earlier for any $\cA$, $x_{1} =0$. Thus, as a function of $\cA$'s action $x_2$ at time $t = 2$,  the cost of \(\cA\)  is
\begin{align}\nn
	C_{\cA}
	&= \frac{\mu}{2}\theta ^ 2 + \frac{\mu}{2}x_{2} ^ 2 +\abs{x_{2}} \ge \frac{\mu}{2}\theta^2, \nn
\end{align}
where the last inequality follows since $\frac{\mu}{2}x_{2} ^ 2 +\abs{x_{2}}\ge 0$ irrespective of the choice of $x_2$.
For the considered choice of \(f_{1}\), \(f_{2}\), the \(\opt\) solves the following optimization problem:\begin{align}\nn
	C_{\opt} &= \min_{x_{1}, x_{2}} \big(\frac{\mu}{2}(x_{1} - \theta) ^ 2 + \frac{\mu}{2}x_{2} ^ 2 + \abs{x_{1}} + \abs{x_{1} - x_{2}}\big), \\
	&\stackrel{(a)}\leq \min_{x_{1}} \big(\frac{\mu}{2}(x_{1} - \theta) ^ 2 + 2\abs{x_{1}}\big), \nn \\
	&\stackrel{(b)}\leq \min_{x_{1} \geq 0} \big(\frac{\mu}{2}(x_{1}- \theta) ^2 + 2x_{1}\big) = 2\big(\theta - \frac{1}{\mu}\big),\nn
\end{align}
where \((a)\) follows by setting \(x_{2} = 0\) in the \(\opt\)'s optimization problem. Further, since \(\theta > \frac{1}{\mu}\), we have $\min_{x_{1} \geq 0} \big(\frac{\mu}{2}(x_{1}- \theta) ^2 + 2x_{1}\big) = 2\big(\theta - \frac{1}{\mu}\big)$ in \((b)\). Therefore, the competitive ratio of \(\cA\) can be lower bounded by \begin{align*}
	\text{cr}_{\cA} \ge \frac{\mu \theta ^ 2}{4\big(\theta - \frac{1}{\mu}\big)}.
\end{align*}
The condition \(\theta < \frac{\sqrt{129} - 9}{2\mu}\) implies \begin{align}\nn
	\theta ^ 2 + \frac{9}{\mu} \theta -\frac{12}{\mu ^ 2} \leq 0 \iff \frac{\mu\theta ^ 2}{4\big(\theta - \frac{1}{\mu}\big)} \ge \frac{1}{8} \big(2\mu \theta + \frac{12}{\mu\theta} + 3 \big).
\end{align}
For the chosen \(f_{1}, f_{2}\), we have \(\cP_{T} ^ \star = \theta, \cP_{2, T}^ \star = \theta ^ 2,\) and \(\abs{x_{1}^\star} = \theta\). Therefore, the quantity in \eqref{lb_linear_any_A} is equal to \begin{align*} \mu(\cP_{T}^\star + \abs{x_{1}^\star}) + \frac{3}{2} \frac{\big(\cP_{T}^\star + \abs{x_{1}^\star}\big) ^ 2}{\cP_{2, T}^\star + \abs{x_{1}^\star} ^ 2} + \frac{12}{\mu} \frac{\cP_{T}^\star + \abs{x_{1}^\star}}{\cP_{2, T}^\star + \abs{x_{1}^\star} ^ 2} = 2\mu \theta + \frac{12}{\mu\theta} + 3.\end{align*}
Therefore, for the considered problem instance, we have \begin{align*}
	\text{cr}_{\cA} \ge \frac{1}{8}\bigg(\mu(\cP_{T}^\star + \abs{x_{1}^\star}) + \frac{3}{2} \frac{\big(\cP_{T}^\star + \abs{x_{1}^\star}\big) ^ 2}{\cP_{2, T}^\star + \abs{x_{1}^\star} ^ 2} + \frac{12}{\mu} \frac{\cP_{T}^\star + \abs{x_{1}^\star}}{\cP_{2, T}^\star + \abs{x_{1}^\star} ^ 2}\bigg),
\end{align*}
which completes the proof.

\end{document}